%% file: main.tex
\documentclass[9pt, final, journal,a4paper,twocolumn]{IEEEtran}

\usepackage{hyperref}
\usepackage{listings}
\usepackage{placeins}
\usepackage{subfig}
\usepackage{amsmath,amssymb}
\usepackage{latexsym}
\usepackage{xcolor}
\usepackage{engord}

\usepackage{graphicx} \graphicspath{{figures/}}

\makeatletter
\newcounter{@medas@demo@cnt}
\def\newdemo#1{
    \refstepcounter{@medas@demo@cnt}%
    \subsection*{Case Study \arabic{@medas@demo@cnt}:~#1}%
    \edef\@currentlabelname{#1}%
    \setcounter{subsubsection}{0}
}
\makeatother

\begin{document}

    \title{MeDaS: An open-source platform as service to help break the walls between medicine and informatics}

    \DeclareRobustCommand*{\IEEEauthorrefmark}[1]{%
    \raisebox{0pt}[0pt][0pt]{\textsuperscript{\footnotesize\ensuremath{#1}}}}
    \author{%
    \IEEEauthorblockN{%
        Liang~Zhang\IEEEauthorrefmark{1}\thanks{Corresponding author: Liang Zhang (Email: \href{mailto:liangzhang@xidian.edu.cn}{liangzhang@xidian.edu.cn}}, Member, IEEE,
        Johann Li\IEEEauthorrefmark{1},
        Ping Li\IEEEauthorrefmark{2},
        Xiaoyuan Lu\IEEEauthorrefmark{2},
        Peiyi Shen\IEEEauthorrefmark{1},
        Guangming Zhu\IEEEauthorrefmark{1},
        Syed Afaq Shah\IEEEauthorrefmark{3},
        Mohammed Bennarmoun\IEEEauthorrefmark{4},
        Kun Qian\IEEEauthorrefmark{5}, Member, IEEE, and
        Bj\"orn W.\ Schuller\IEEEauthorrefmark{6,7}, Fellow, IEEE
    }\\\vspace*{1em}
    \IEEEauthorblockA{\IEEEauthorrefmark{1} Embedded Technology \& Vision Processing Research Center, School of Computer and Technology, Xidian University, Xi’an, China}\\
    \IEEEauthorblockA{\IEEEauthorrefmark{2} Shanghai BNC, Shanghai, China}\\
    \IEEEauthorblockA{\IEEEauthorrefmark{3} College of Science, Health, Enginering and Education, Murdoch University, Australia}\\
    \IEEEauthorblockA{\IEEEauthorrefmark{4} School of Computer Science and Software Engineering, The University of Western Australia, Australia}\\
    \IEEEauthorblockA{\IEEEauthorrefmark{5} Educational Physiology Laboratory, The University of Tokyo, Japan}\\
    \IEEEauthorblockA{\IEEEauthorrefmark{6} GLAM - Group on Language, Audio \& Music, Imperial College London, UK}\\
    \IEEEauthorblockA{\IEEEauthorrefmark{7} Chair of Embedded Intelligence for Health Care and Wellbeing, University of Augsburg, German}
    }

	\maketitle

    \input{abstract}
    \input{introduction}
    \input{relatedwork}
    \input{rinv}

    \input{pipeline}

    \input{system}
    \input{application}
    \input{discussions}
    \input{summary}

    \bibliographystyle{IEEEtran}
    \bibliography{refs/references-1b}

\end{document}

%% file: abstract.tex

\begin{abstract}
    In the past decade, deep learning (DL) has achieved unprecedented success in numerous fields including computer vision, natural language processing, and healthcare. In particular, DL is experiencing an increasing development in applications for advanced medical image analysis in terms of analysis, segmentation, classification, and furthermore. On the one hand, tremendous needs that leverage the power of DL for medical image analysis are arising from the research community of a medical, clinical, and informatics background to jointly share their expertise, knowledge, skills, and experience. On the other hand, barriers between disciplines are on the road for them often hampering a full and efficient collaboration. To this end, we propose our novel open-source platform, i.\,e., \textsc{MeDaS}--the MeDical open-source platform as Service. To the best of our knowledge, MeDaS is the first open-source platform proving a collaborative and interactive service for researchers from a medical background easily using DL related toolkits, and at the same time for scientists or engineers from information sciences to understand the medical knowledge side. Based on a series of toolkits and utilities from the idea of RINV (Rapid Implementation aNd Verification), our proposed \textsc{MeDaS} platform can implement pre-processing, post-processing, augmentation, visualization, and other phases needed in medical image analysis. Five tasks including the subjects of lung, liver, brain, chest, and pathology, are validated and demonstrated to be efficiently realisable by using  \textsc{MeDaS}.
\end{abstract}


%% file: introduction.tex

\section{Introduction}
\label{sec:intro}

Deep learning is the present cutting-edge technique in computer vision, medical image analysis, and several other areas. Thanks to its power, researchers can use a regular pipeline to process and analyze images and obtain excellent results with the aid of deep learning.
For instance, there are a lot of recent studies that apply deep learning   \cite{Litjens2017,Kamnitsas2016,Dolz2018,Yi2019,Wang2020,Swiderska-Chadaj2019}.
However, most researchers, who use deep learning in their research on medical image-related tasks, are professionals in computer science, and not medicine. Due to the often present lack of computer related knowledge, it is hard for medical researchers to understand and apply deep learning in their research individually for tasks such as tumor segmentation and nuclei classification. As to computer science researchers, they cannot fully analyze their results without the help of medical researchers. This present gap between computer science and the medical field creates a bottleneck for the use of deep learning in medical image analysis.

Programming is a skill. Programmers  design  programs with a series of instructions to operate hardware. However, directly operating hardware with instructions is very difficult for most people. Therefore, one important concept in computer-related areas is ``code reuse''. The experts in one area create frameworks, libraries, for others to  implement their programs easily.
TensorFlow (\cite{Abadi2016}), ITK (\cite{Ibanez2003}), and OpenCV (\cite{Paszke2017}) are typical examples from the target domain of deep learning and image analysis to help researchers simplify their programs.

In medical areas, ITK~(\cite{Ibanez2003}), ANTs~(\cite{Avants2011}), FSL~(\cite{Jenkinson2012}), Deep Neuro~(\cite{Beers2018}), and NiftyNet~(\cite{Gibson2018}) are the prevalent toolkits, libraries, and frameworks to help medical researchers build programs to analyze medical images and data.
These tools, libraries, and frameworks can help them to register different images, process, visualize, and analyze them.
However, it still requires a significant level of programming skills from interested medical researchers who want to apply such toolkits, libraries, and frameworks.

While using computers to solve a problem, the approaches can be divided into three levels. The \textit{\textbf{first level}} is to execute the task by all yourself; the \textit{\textbf{second level}} is to combine other libraries via programming; the \textit{\textbf{third level}} is  to use out-of-the-box software and interact via a user interface. Meanwhile, the most toolkits, libraries, and frameworks in the target domain of interest here provide only the first and the second levels, which still require programming skills. The \textit{\textbf{first level}} and \textit{\textbf{second level}} require expert programming skills and limit the access of those who have not majored in computer science. For example, the popular deep learning frameworks only provide clear interfaces of the \textit{\textbf{second level}}, and users are required to code when attempting to use deep learning for  medical data analysis. This creates a challenging situation for these researchers.

As most deep learning frameworks, such as TensorFlow and PyTorch, only provide APIs (programming), the application of deep learning becomes a challenging problem for medical researchers, who are not familiar with programming.
However, when we take a closer look at the most use-cases of deep learning-based medical image analysis, one easily sees that pre-processing, augmentation, neural networks, post-processing, visualization, augmentation, and debugging is the commonly used pipeline. Therefore,  users with a purely or mostly medical background would not need to implement these algorithms, but rather reuse them. Furthermore, those researchers could simply combine these tools and make up their models without programming when applying deep learning in their studies with visual programming.

Nevertheless, all  these frameworks and toolkits are not integrated as a system. Researchers need to assemble their program from here and there one by one with their programming skills, unlike  out-of-the-box tools, such as Microsoft Excel and IBM's SPSS. In order to help medical researchers build their deep learning models easily, a \textbf{MEDical open-source platform As Service}(\textbf{MeDaS}) is proposed in the oncoming.

The main idea of MeDaS is to provide a scalable platform as a service and integrating a set of tools ~to cover the implementation of deep learning models for medical image analysis. Moreover, MeDaS not only provides commonly used tools, functions, and modules used in deep learning, but can also help researchers to manage their computing resources and refine their models.

The remainder of this article is organized as follows.
\textbf{Section \ref{sec:rew}} introduces the related work on software, medical, Docker, and other technologies.
\textbf{Section \ref{sec:rinv}} expounds our main idea of rapidly implementation and verification, i.e., \textbf{RINV}. \textbf{Section \ref{sec:pipeline}} discusses the basic components that MeDaS provides to users to design and implement their algorithms and models. \textbf{Section \ref{sec:sys}} introduces the utilities that MeDaS provides to simplify programming, management, and refining.
\textbf{Section \ref{sec:app}} introduces several case studies of MeDaS, including \textit{lung property classification}, \textit{liver contour segmentation}, \textit{multi-organ segmentation}, \textit{Alzheimer's Disease classification}, \textit{pulmonary nodule detection}, and \textit{nuclei segmentation}. Finally, Section \ref{sec:dscs} provides a discussion of open questions, and Section \ref{sec:sum} concludes the paper.

%% file: relatedwork.tex

\section{Related Work}
\label{sec:rew}

In this section, we will discuss the related toolkits and software for medical image analysis, the deep learning frameworks used in most relevant works, and other technologies used in or related to MeDaS.

\subsection{Medical Toolkits}

\paragraph{ANTs} Advanced Neuroimaging Tools\cite{Avants2011} is a toolkit for brain images, and provides functions to visualize, process, and analyze the multi-modal image, and others.

\paragraph{FreeSurfer} FreeSurfer \cite{Fischl2012} is an open-source toolkit for processing and analyzing MRI images, which includes functions about skull stripping, image registration, subcortical segmentation, cortical surface reconstruction, cortical segmentation, cortical thickness estimation, longitudinal processing, fMRI Analysis, tractography, and GUI-based visualization.

\paragraph{ITK} Insight Segmentation and Registration Toolkit  \cite{Lowekamp2013}  is the most popular toolkit widely used in medical image analysis. The functions provided by ITK include basic operations of medical images, visualization, preprocessing, registration, and segmentation. It is implemented with C++, and offers template and bindings for Python, Java, and other languages.

\subsection{Deep Learning-based Medical Toolkits}
\paragraph{DeepNeuro} DeepNeuro \cite{Beers2018} is an open-source toolkit with deep learning pipelines and applications, which provides open-box-fee pipelines and applications. It aims at medical image analysis with deep learning.

\paragraph{MIScnn} Medical Image Segmentation with Convolutional Neural Networks \cite{Muller2019}, which was released  recently, targets medical image segmentation based on Convolutional Neural Networks and Deep Learning, and provides pipelines and programming-based user interface to help users to create their dedicated models.

\paragraph{NiftyNet} NiftyNet \cite{Gibson2018}, is another open-source toolkit, similar to DeepNeuro, which provides a series of components such as dataset splitting,  data augmentation, data pre- and post-processing, a pre-designed network, and evaluation metrics.  NiftyNet aims at medical image analysis with deep learning.

\subsection{Deep Learning Frameworks}
Many researchers use deep learning methods to analyze medical images, and they rely on frameworks to ease their research. The framework requires the users to implement algorithms and run them on a GPU by themselves; thus, the researchers are forced to spend a lot of time on testing and implementation.

\paragraph{Caffe} Caffe \cite{Jia2014} created the Caffe framework, which is an abbreviation for Convolutional Architecture for Fast Feature Embedding. It provides a useful open-source deep learning framework. Caffe filled the gap between different devices and platforms.

\paragraph{PyTorch} Facebook released Torch -- a scientific computing framework. It widely supports machine learning algorithms on the GPU. A few years later, Facebook released another deep learning framework, named PyTorch, \cite{Steiner2019, Paszke2017}, which puts Python first, and is one of the most popular deep learning frameworks for researchers.

\paragraph{TensorFlow} Google released a deep learning framework named TensorFlow \cite{Abadi2016}, aimed at tensor-based deep learning. TensorFlow is based on dataflow graphs and can run on different devices, including CPU, GPU, and Google's TPU. The platforms can vary from personal computing to server clusters. TensorFlow is widely used for research and in industry. Google has also open-sourced a number of tools for TensorFlow, such as TensorBoard.

\subsection{Docker and Visual Programming}

Docker \cite{docker} is a kind of container platform, and also is an industrial level resource management solution. Docker takes on the management task of computing resources, which frees its users to focus on their research. It allows containers launched in a short time, and also allows the mass of applications to run on the host and keep the host and containers ``clean''.

NVIDIA released nvidia-docker\cite{nvidia-docker} in 2015, which makes it possible to use a CUDA-enabled GPU in Docker containers. In this way, one can use the GPU to accelerate one's algorithms in Docker.

Kubernetes\cite{k8s} is one of the most famous Docker cluster management software pieces, which can save one from managing a lot of workstations or servers. Users can simply upload their tasks, run the tasks on a machine, and supervise their tasks on a web-based user interface.

Visual programming allows users to create programs by manipulating program pipelines graphically or drag and drop elements, such as Unreal Engine's Blueprints Visual Scripting \cite{ue-bvs} and Scratch \cite{Maloney2010}. This allows naive programmers or the researchers not familiar with programming to build  machine learning models quickly by dragging and dropping pipelines. With visual programming, researchers can monitor their algorithms or the mathematical models.

%% file: rinv.tex

\begin{figure}
	\centering
	\includegraphics[width=\linewidth]{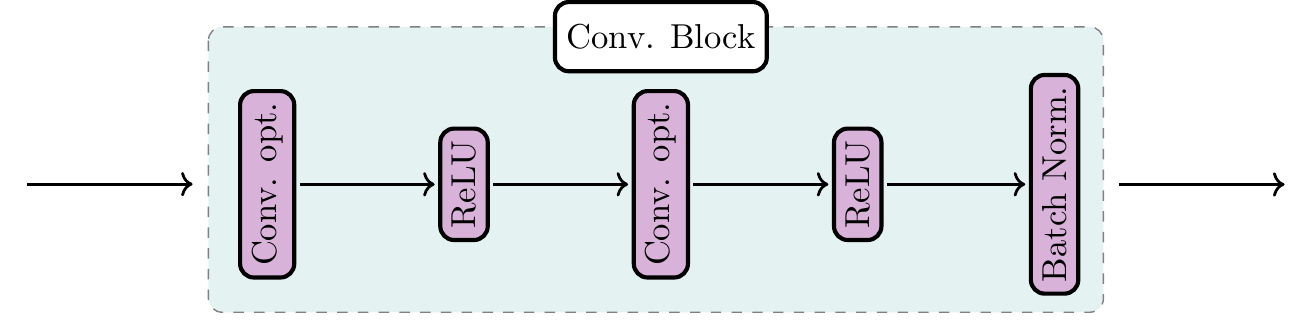}
	\caption{The ``convolution'' block is combined with convolution layers, ReLU active functions, and batch normalization. A deep learning model is also combined with other modules. These form pipelines.}
	\label{fig:brick-combine}
\end{figure}

\begin{figure}
	\centering
	\includegraphics[width=0.9\linewidth]{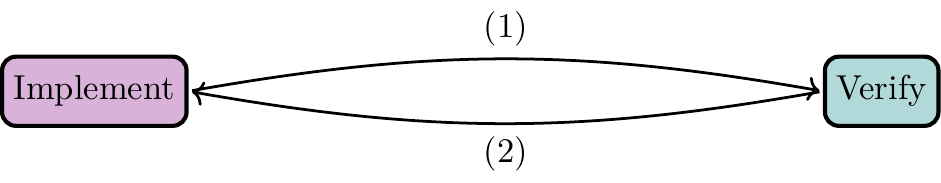}
	\caption{The workflow is a circle of implementation and verification. For example, in the development of the algorithm, the verification is followed by implementation to assure the algorithm works as planned. The implementation is also followed by verification, to fix errors and improve the performance. After several iterations, the development is finished.}
	\label{fig:implement-workflow}
\end{figure}

\section{Rapid Implementation and Verification}
\label{sec:rinv}

The naive motivation behind MeDaS is to make the application of deep learning easier for medical researchers in their research. However, fact is that deep learning requires programming skills, which can -- as laid out -- be challenging for medical researchers without according background knowledge. That makes using frameworks such as TensorFlow and PyTorch difficulty, but there is another way to help create models and algorithms with deep learning. For medical researchers, deep learning is mainly a tool. Hence, ideally, it is a good idea that they are provided simply with high-level user-friendly software to implement and verify their models and algorithms rapidly.

The idea to implement and verify a model  is called ``\textbf{Rapid Implementation aNd Verification}''(\textbf{RINV}). RINV aims at the workflow from the formula or algorithm from draft stage to the final program and results. \textbf{Based on this idea, MeDaS provides tools and utilities to help focus on one's model, and simplify the implementation and verification.} The tools and utilities will be introduced in Section \ref{sec:pipeline} and Section \ref{sec:sys}.

The medical image analysis is a significant  computer vision task, but deep learning, which is commonly used in computer vision, is not favored by most medical researchers and doctors, because they know too little about deep learning, especially programming. Just like a medical researcher should not have to build a CT scanner before he or she wants to scan, they should not be required to spend unnecessary time on the implementation of deep learning before using it, either.
Though there is no need to implement deep learning algorithms from cover to cover, programming is still a difficult thing.

Most of the algorithms and mathematical models are the combination of sub-algorithms, sub-pipelines, or subroutines, as Fig.\ \ref{fig:brick-combine} shows. Not only in medical  areas, but also in  computer science, physics, and other disciplines, the computer programs are created by combining subroutines. Meanwhile, each component comes with a workload shared between implementation and verification as can be seen in Fig.\ \ref{fig:implement-workflow} (path (1) and (2)).

When implementing an algorithm or model for diagnosis, recognition, or segmentation, it can be a waste of time to focus on the implementation and verification of the algorithms. Instead, medical researchers should spend time on designing the model itself.
Focusing on implementing and verifying a diagnosis, recognition, and segmentation system is just the half opposite of what researchers might desire to focus upon. They should be given the opportunity to focus on designing models and algorithms.

\begin{figure}[t!]
	\centering
	\includegraphics[width=\linewidth]{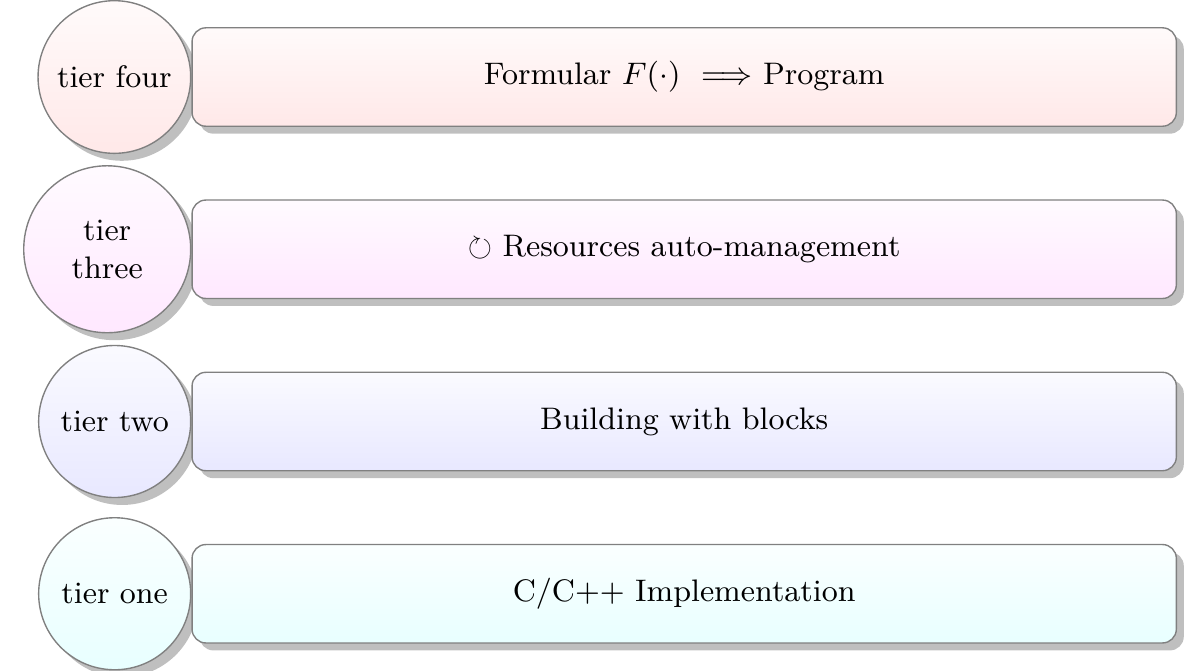}
	\caption{There are four tiers of deep learning model development. Tier one is the implementation with C/C++ and assembly, such as for  cuDNN(\cite{Chetlur2014}). The next tier is the combination of the basic blocks. Tier three includes the management of resources to help users focus on the model itself. Tier four aims at converting the model to program directly, meaning the implementation and verification is automatically completed by the software.}
	\label{fig:Rinv-level}
\end{figure}

There are a lot of steps involved in converting the model to program even a basic system. The process of transforming from a model or formula to the program can be split into four tiers, as shown in Fig.\ \ref{fig:Rinv-level}.
At \textit{\textbf{tier one}}, researchers need to do everything by themselves. They need to implement and verify the low-level algorithms with C++ and assembly, convert mathematical formula to a program, make sure the program runs on the right device, manage computing resources, visualize results, and so on.
At \textit{\textbf{tier two}}, researchers can use naive algorithm toolkits to implement the complex program, but they still need to manage the device resources by themselves.
At \textit{\textbf{tier three}}, the management of computing resources will be scheduled automatically or non-manually.
\textbf{The \textit{tier four} is inputting mathematical formula and outputting results.}

Tier four is a moonshot, but still a utopian design. However, researchers mostly prefer tier four, which outputs the result with a given  model and without any coding. Our aim for MeDaS is to realize functions of tier three, which can provide efficient algorithms for users to implement their models and algorithms, and help them manage their resources efficiently.

%% file: pipeline.tex

\begin{figure}[t!]
	\centering
	\includegraphics[height=0.5\paperheight]{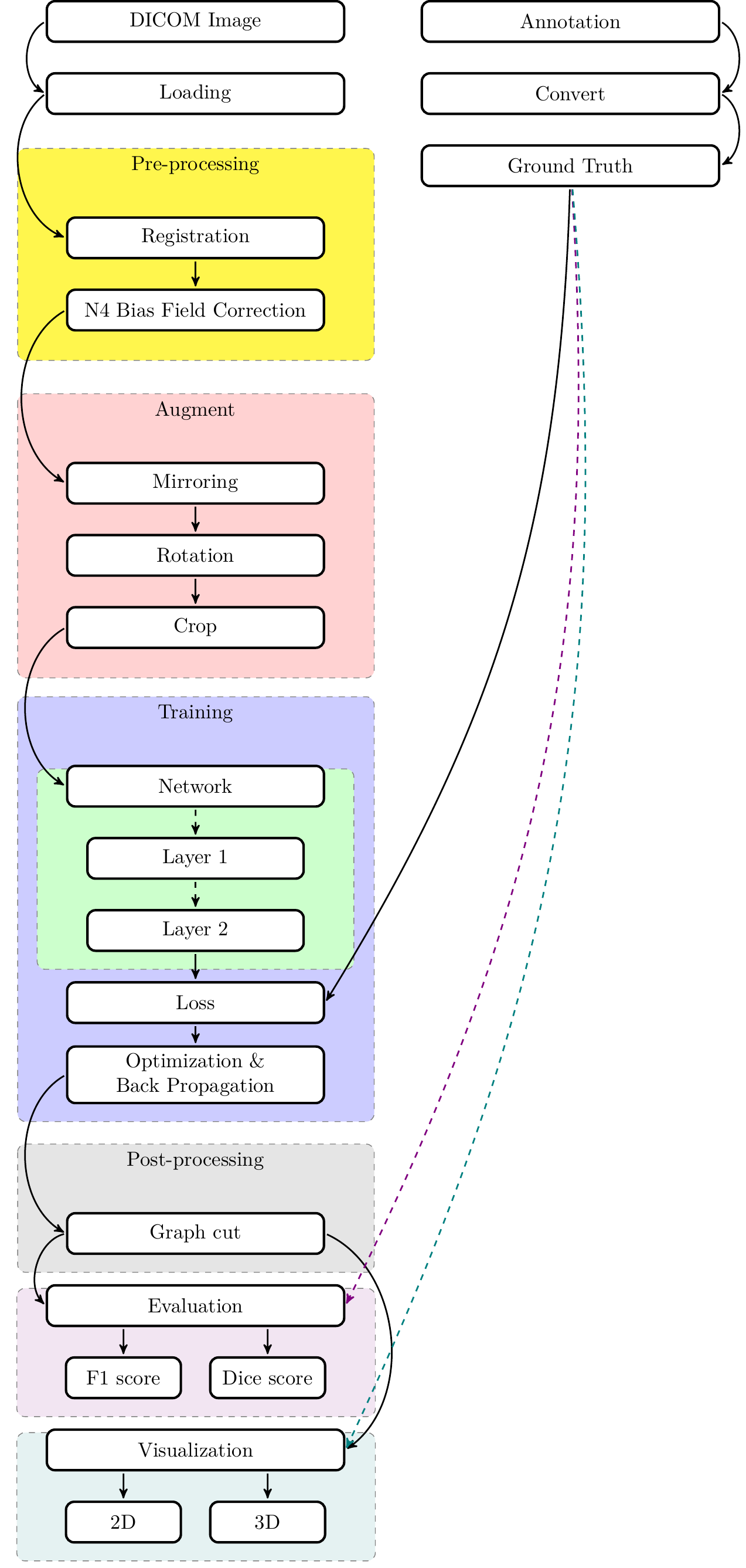}
	\caption{The general flow of an application of deep learning for MRI image analysis. The flow shows pipelines and components of pre-processing, post-processing, augmentation, evaluation, visualization, and a neural network. The pre-processing includes registration and N4 bias field correction. The augmentation contains mirroring, rotating, and cropping. At the same time, annotations are processed for training to be used as the ground truth. The network is composed of a series of layers, such as convolution, ReLU function, and further more. In each iteration, the loss is calculated  and used for the optimization algorithm, which `searches' for the best parameters of the model. The post-processing uses graph-cutting and handles the data predicted by the neural network and renders the result optimised. The evaluation layers exploit the ground truth and the prediction result to evaluate the performance of the model, while the visualization layers are designed for debugging and visualization.}
	\label{fig:deep-learning-work-flow}
\end{figure}

\section{Core: Tools of Deep Learning}
\label{sec:pipeline}

Similar to NiftyNet and MIScnn -- and as laid out -- MeDaS provides a series of tools to allow users to combine them to create algorithms and models. With the idea of ``Rapid Implement and Verification'', MeDaS also employs visualization for a programming interface. The tools are discussed in this section, and the whole architecture of MeDaS is given in Fig.\ \ref{fig:vispg} and Section \ref{sec:sys}.

The tools, provided by MeDaS, can help researchers enhance their algorithms and models, and a neural network is provided as a tool. Therefore, the assembly of tools is a most simple way to implement complex algorithms.

For medical imaging-related fields, the workflow of processing of medical images is relatively fixed. For both, traditional methods, such as PCA and graph cut, and recent ones, for example, deep learning-based methods, the workflow usually includes:
\begin{itemize}
	\item Dataset management
	\item Pre-processing
	\item Kernel algorithm
	\item Post-processing
	\item Visualization and evaluation.
\end{itemize}

After analyzing the pipeline of deep learning from our and others' research, \cite{Henschel2019,Skibbe2019,Zhang2019,Rajan2019,Rajchl2018,Zhang2017,Crankshaw2018,Lee2015},
we found that the pipeline in these contributions share a similar view. \textbf{Pre-processing} (\cite{Yao2017, Jeyavathana2016,Ferdouse2011,Ogiela2008,Kamnitsas2016}), \textbf{augmentation} (\cite{Kamnitsas2016,Litjens2016,Pereira2016,Shorten2019}), \textbf{post-processing} (\cite{Kamnitsas2016}), \textbf{visualization} (\cite{Yong2012}), \textbf{debugging}, and other pipelines are mostly used.  The importance of these pipelines is obvious. Fig.\ \ref{fig:deep-learning-work-flow} shows the workflow  of the typical deep learning for medical image processing pipelines.

Each step/pipeline has its purpose of processing. Therefore, MeDaS implements a series of tools to meet these requirements, including image preprocessing, post-processing, augmentation, artificial neural network, visualization, and other steps.

\subsection{Pre-processing}
As the name implies, pre-processing is the step before training of the neural networks. It includes feature processing, such as feature extraction, noise reduction, data normalization, modalities' registration, and data processing, such as format conversion, annotation transformation and further more. We implement the necessary tools to help researchers process the data before the training of their models.

There usually exists a strong data bias in medical images. For radiography, such as CT and PET, the images are noisy due to the different pieces of equipment, \cite{S.Rameshkumar2016, S.Senthilraja2014}, different operators, and even different settings.
Therefore, MeDaS implements the basic \textbf{registration tool} and \textbf{N4 bias field correction tool}, \cite{Tustison2010},  to help one process one's data.

For pathology ,  the difference of stain concentration produces different results for the algorithms, \cite{Magee2009, Reinhard2001, Ruifrok2001}. Thus, \textbf{stain normalization tool}, \cite{Vahadane2016}, and \textbf{stain deconvolution} \cite{Ruifrok2001} are applied as the space of stain is not linear and general normalization tool does not work.

Meanwhile, for general purposes, the \textbf{normalization tool}, \textbf{resample tool}, \textbf{rescale tool}, \textbf{mask generating tool}, \textbf{resize tool}, and other tools are implemented to process data.
Furthermore, MeDaS implements serial tools, including \textbf{format conversion tools}, \textbf{annotation conversion tools}, and further more.

\subsection{Augmentation}

The scale of datasets in the concerned medical areas is usually considerably smaller than than in others, \cite{Bilic2019,Setio2016,Kumar2017}. The public medical image datasets generally have 100 to 1000 cases, while  other datasets -- for example, for 3D object detection \cite{Wu2014} -- usually feature  thousands and even millions of data. Therefore, augmentation is necessary to enlarge the size of the dataset.  Medical image datasets `always' lack data, compared to other areas, because data acquisition and annotation takes a lot of time, cost, and manpower.

Augmentation is an efficient method to make models more robust, not only in medical image analysis, but also in other areas.
Augmenting with mirroring, rotating, cropping, format transforming and other methods such as by Generative Adversarial Networks are frequently used. Augmentation diversifies the data making it {``different''} -- which can improve the model performance, \cite{Madani2018,Goodfellow2015}.
The key to augmentation is that the distribution of data is expanded such that it leads to increased robustness of the model.

MeDaS provides general transformation tools ~\textbf{Gaussian random noise}, \textbf{rescaling} tools {~and other tools. The former uses noise to enhance the robustness of the model, while the other two desensitize the noise of the scale and the bias by re-sampling and transforming the distribution of the data.

\subsection{Artificial Neural Network}

The neural network is the most important part in deep learning, simply, as  deep learning is essentially based on a deep neural network. MeDaS provides several tools to integrate different types of neural networks. These tools can be used for training or inferring. Meanwhile, MeDaS plans to integrate a  neural architecture search, which aims at automatically designing neural networks for specific tasks.

Network (model) training is a fixed workflow which includes forward propagation, loss calculation, and backward propagation \cite{LeCun1998a}.
The neural network is modularized.
The network is built by connecting ``blocks'' such as ``max-pooling layer'', ''convolution layer'', ``fully connected layer'', ``ResBlock'', ``Dense Block'', and so on \cite{LeCun1998b, Huang2016,He2015,Simonyan2014,Krizhevsky2012,Szegedy2016,Ronneberger2015}. Loss function influences the search in the parametric space, the different loss functions meet the different tasks.
As the neural network is intended to be applied merely as a tool by medical researchers,  they are considered as users and not developers. Therefore, the tools with pre-designed networks can be the best choice and can meet needs of researchers that want to focus on the application side of matters.

Since only a few neural networks have achieved significant success for many medical image analysis tasks, MeDaS implements those network as tools for segmentation, classification, and other tasks. For instance, the \textbf{3D Mask RCNN} (\cite{He2017}) and \textbf{3D Dual-Path Net} (\cite{Chen2017}) are integrated for the detection and classification tasks on radiography images. The \textbf{U-Net} (\cite{Ronneberger2015}) and \textbf{V-Net} (\cite{Milletari2016}) are integrated for the segmentation task, besides the U-Net also being available to be used in classification tasks. The other similar neural networks are also integrated for segmentation and classification tasks.

\subsection{Post-processing}
Post-processing is a strategy that can improve the result. For segmentation tasks, post-processing can make predictions more ``smooth''.
For example, \cite{Graham2018} employed an FCN-based neural network, which is simpler to UNet and VNet, but achieves better performance compared to the case study \ref{case:ns}, which applies UNet-based networks. The key to its success is the post-processing step.  \cite{Graham2018} use ``horizontal and vertical gradient maps'', ``energy landscape'', and other features as the post-processing, e.g., the \textbf{watershed algorithm}.

MeDaS integrates many post-processing tools.  A \textbf{Conditional Random Field} (\cite{Chen2018}), \textbf{Graph Cut} (\cite{Jimenez-Carretero2019}), and other traditional algorithms can be used as post-processing to optimize the results of a neural network.

In a few cases, the output of the neural network is a  probability or a probability map. The tools, such as \textbf{binary normalization}, can
be used for the classification and segmentation tasks, which will reach better results compared to a simple threshold.

\subsection{Visualization}

Generally speaking, the visualization can be categorized into result visualization, metric visualization, and analysis visualization.

The result and analysis visualization show the result of the final inference, which keeps important links between the model and clinic side  \cite{Jamaludin2017,Hohman2019,Zhang2018,Yong2012,Liu2019}.
The results of algorithms, such as segmentation and classification, are data-based -- this is hard to be shown directly, as it is not a color-based image. A well-designed tool of visualization can help users present and analyze their work corresponding with the clinical aspects.

For metric visualization, MeDaS implements tools to visualize the metric as an image, for example, a \textbf{loss visualization} tool.
For result and analysis visualization, MeDaS implements a series of tools for many kinds of tasks.
The \textbf{segmentation visualization}, \textbf{organ visualization}, \textbf{point cloud-based pulmonary nodule visualization}, and other visualization tools are implemented for visualization. MeDaS also implements analysis visualization tools, such as \textbf{sensitivity analysis} tools to help researchers to analyze their results.

\subsection{Others}

MeDaS also includes other kinds of tools, such as for dataset management and analysis tools.
The former controls the dataset used in neural network training, while the latter analyzes the results.

The dataset management tool aims at the management of the dataset. For example, if one wants to split one's data into a training set and a testing set, one can use the \textbf{dataset split} tool to carry this step out.

The \textbf{result analysis} and \textbf{metric} tools can help analyze  results of the analysis tools.

%% file: system.tex

\begin{figure}[t!]
	\centering
	\includegraphics[width=\linewidth]{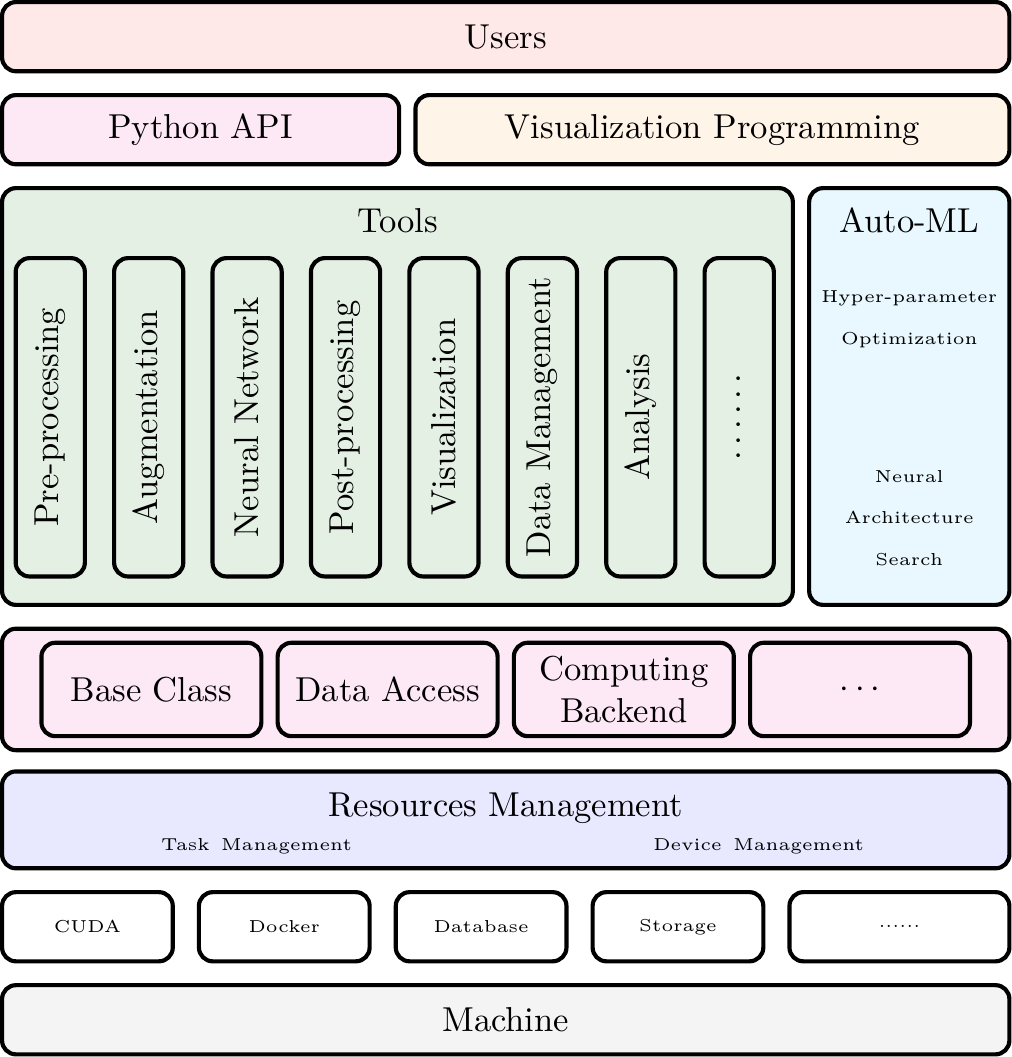}
	\caption{The general architecture of MeDaS: From the bottom (machine) to the top (user). The user can operate MeDaS via a Python API (Section\ref{sec:sys:py}) or a visualization programming interface (Section\ref{sec:sys:vp}) to operate tools (Section\ref{sec:pipeline}), auto-machine learning (Section\ref{sec:sys:automl}), resources management (Section\ref{sec:sys:rcmg}), and other components.}
	\label{fig:vispg}
\end{figure}

\section{Architecture of MeDaS}
\label{sec:sys}

MeDaS not only includes the core - tools, as shown in Fig.\ \ref{fig:vispg}, but also other parts.Different from traditional toolkit, MeDaS is a kind of system, and not a collection of tools.
The traditional toolkits and frameworks can only be applied via programming, while MeDaS provides a visualization programming approach to help researchers intuitively and easily use our MeDaS. In the following, we will discuss the visualization-based programming, auto-machine learning, Python APIs, and resources management features of MeDaS.

Fig.\ \ref{fig:vispg} shows the architecture of MeDaS. From the bottom to top, the figure depicts each component of MeDaS, including the named features: Visualization programming, auto-machine learning, python API, and resource management. The users can interact with MeDaS via Python API or visualization programming. The operation with tools is directly concerned with the Python API, while the visualization provides more functions integrated in MeDaS, such as auto-machine learning. The resources management is a part of MeDaS, but MeDaS does not provide any programming APIs for it. The resources management controls the tasks scheduling and device allocation, which directly interact with the machine.

\subsection{Visualization Programming}
\label{sec:sys:vp}

The original interface of a computer is teletypewriter-based. Later, scientists invented a terminal based on CRT. Until the invention of the Graphical User Interface (GUI), there was no way to graphically interact with software for non-professionals. The computer has its own rules. Software developmenter convert instructions from the ``human rules'' to ``computer rules'', and that is called ``implementation''. GUI-based sotfware can efficiently help non-professionals to translate their ideas from ``human rules'' to ``computer rules'' and to execute them.

A GUI is usually considerably more intuitive than a Command Line Interface (CLI) or any text-based interface -- especially for the ones not or less familiar with computers. If well designed, it can render the operation of tools and visualizing results more accessible and efficient for its users.

As shown in Fig.\ \ref{fig:vispg}, visualization programming is used to interact with users directly. They use such an interface to operate the MeDaS via dragging, dropping, and connecting. Moreover, visualization programming is accessible directly online via a web site, and users do not need to install any client, but a web browser suffices.

\subsection{Auto-Machine Learning}
\label{sec:sys:automl}

Designing and optimization of the deep neural network are the keys to success in the current state-of-the-art medical imaging approach. However, the design and refinement of the model are not trivial.
Therefore, MeDaS integrates  auto-machine learning utilities, as shown in Fig.\ \ref{fig:vispg}.

Generally, the parameter $\theta$ of a deep learning model $f(x; \theta)$ can be optimized by gradient descent, while the hyper-parameter needs to be optimized manually, and the model needs to be designed by hand as well. Furthermore, researchers need to design their own networks or choose from a large number of out-of-the-box networks. This adds to the extra workload of users. With that on mind, MeDaS employs automated hyper-parameter optimization and neural architecture search.

%
%

\begin{figure}[t!]
	\centering
	\subfloat{\includegraphics[width=\linewidth]{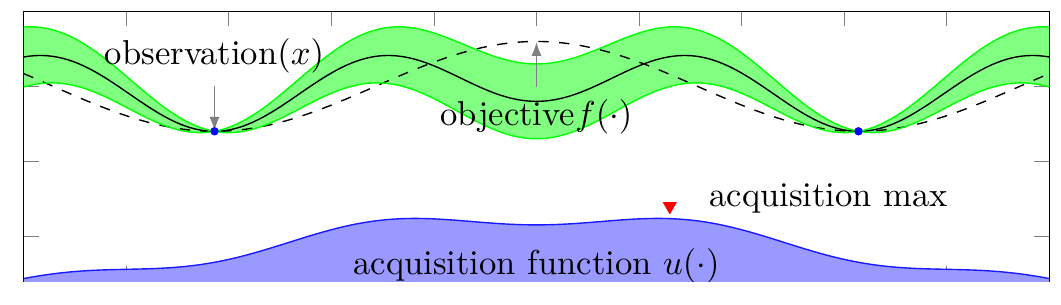}} \\
	\subfloat{\includegraphics[width=\linewidth]{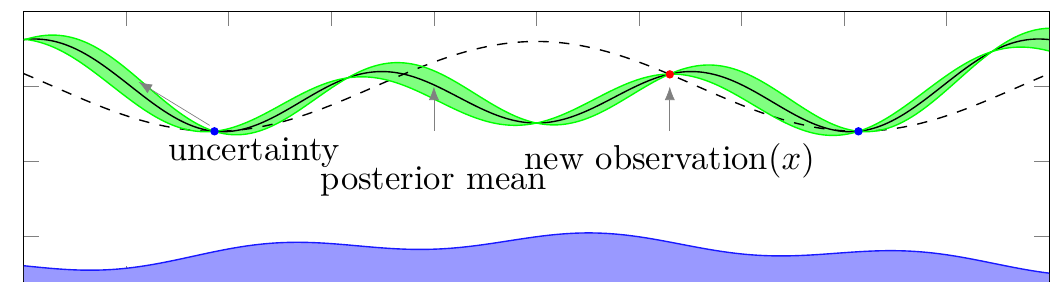}} \\
	\caption{Principle of the provided Bayesian-based hyper-parameter auto-search. The \textbf{above figure} shows the `prediction' of $t=t_1$. The blue points show the observation $x$; the \textbf{black line} presents the posterior mean of the prediction; the dashed line is the objective function $f(\cdot)$; the \textbf{green area} represents the possible functions, while the \textbf{blue area} is the acquisition function $u(\cdot)$. The \textbf{maximum point} of $u(\cdot)$ is the next point of the hyper-parameter to be optimized.
	We use a set of the sine function to explain how Bayesian optimization searches the hyper-parameter. The key idea of Bayesian optimization is the iterative repetition of fitting and search. The methods, such as a Gaussian process and a regression random forest, are employed for fitting the data $(x, y)$, where $x$ denotes the hyper-parameter, and $y$ denotes the performance of the model.
	The acquisition function, such as Expected Improvement and Upper Confidence Bound, is employed for searching the next best $x$ of the model.}
	\label{fig:bayesian-opt-func}
\end{figure}

When we optimize the hyper-parameter $\Theta$ of the model $f(x; \theta)$, we actually need to optimize another model $F(\Theta; f)$, which represents the best value of the function $f$ with the hyper-parameter $\Theta$, to obtain the optimal hyper-parameters.
For optimization $\arg\max F(\Theta; f)$, it is hard to deduce the analytical formula of $F(\cdot)$; hence, we use a set of functions $\{\mathcal{F}\}$ to estimate the distribution of $F(\cdot)$ as Fig.\ \ref{fig:bayesian-opt-func} shows. After training the original model and getting the hyper-parameter result of $F(\cdot)$,
we can remove the functions which do not fit the result. Then, we get a subset $\{\mathcal{F}\}_i$. After several iterations, the distribution of $\{\mathcal{F}\}$
approximates the final one. Ultimately, we can obtain an approximation of the optimal hyper-parameters.

\subsubsection{Neural Architecture Search}

The rule to design the neural network cannot be expressed with a formula or any other mathematical approach. Thus, algorithms to search for the best architecture of a neural network were suggested \cite{Pham2018, Zoph2016}.
The skillful design of a neural network can be time consuming and difficult and requires expertise. However, an automated neural architecture search can help design networks that perform well. Together, neural architecture search and hyper-parameter optimization can help to overcome the difficulty of manual neural network design and refinement.
MeDaS employs it to help researchers, and it provides a system based on the idea published in \cite{Pham2018} to help its users build the neural network for their specific tasks.

\subsection{Python API}
\label{sec:sys:py}

For most researchers without advanced programming background, visualization programming is the best interface and approach to implement their algorithms. However,  for those who are skilled programmers, the Python API might be better suited than visualization programming. Therefore, MeDaS also supports access via a Python API.

At the programming level, MeDaS designs and implements a ``base class'' as the basic class of the tools and to support visualization programming. The base class handles the tasks of input and output, provides the functions of continuous programming, the computing back-end, unified-data processing, and other utilities. The detail descriptions are presented in the following sub-sections.

\subsubsection{Data, Format, Input and Output}

Each type of medical image can have more than one format. Therefore, the ``base class'' employs SimpleITK and OpenSlide \cite{Goode2013} to handle the different formats of medical images. Furthermore, MeDaS supports loading and saving png-type images (both, a single image, and a series of images) and Numpy objects.

\paragraph{Plug and Slot}

The inputs to a tool might be all kinds of files, numbers, or just a numpy array. Therefore, MeDaS employs ``plug'' and ``slot'' to process these inputs with differentiation and to send them to the kernel function with assimilation.
The plug takes charge of the input processing, while the slot handles the input and pass it to the kernel. The plug will also automatically convert the format of the input. For example, when the input is a string, but the parameter should be a float number, the plug will try to parse the string.

\paragraph{Constructor}

Similar to input, the output can also have many formats. Therefore, MeDaS employs ``constructor'' to process the result of the kernel function. The constructor converts the results to different kinds of formats, including DICOM, NIfTI, and numpy array. The variable simply passes through the variable constructor, while the image constructor saves the tensor to an image file or passes it on to the following modules.

\subsubsection{Computing Backend}

MeDaS employs Numpy, OpenCV, and other libraries to implement algorithms, but not C/C++. The low-level algorithm's implementation is not a high priority, due to the lack of time and manpower. However, there is a reserved ``Computing Back-end''. It is inspired by TensorFlow's design. The implementation of faster CPU versions or other device versions, such as GPU and FPGA, can be added to the system via the ``Computing Back-end'', and can be selected when executing the instance initialization.

\subsubsection{Continuous programming}

Inspired by Either Monad in Haskell, \cite{Radul2001, Marlow2010}, MeDaS implements an abstract class named ``Either'', which aims at the processing results and errors. ``Either'' of MeDaS has two states: success and failure, just like the one in Haskell. The actions of operating a tool are executed one by one. The previous execution should be succeed before the current one has been executed. For example, setting up parameters must be done successfully before calculating.

\subsubsection{Others}

\paragraph{Logging} MeDaS employs a flexible logging system, which accepts outputting to a terminal or another system.
Such a logging system supports users to monitor, diagnose, and debug models flexibly.

\paragraph{Testing suit} MeDaS provides a small kit for testing, by which modules included in MeDaS or by third-parties can be well tested.
At the same time, we employ tools to test MeDaS automatically which is known as continuous integration.

\subsection{Resource Management}
\label{sec:sys:rcmg}

Resource management is important in deep learning, medical image analysis, and several other tasks. Let us discuss a situation:

When a researcher uses one computer with one GPU, the management means execution and termination by the researcher. When two researchers share one computer with a GPU, communication between the two researchers is needed for the scheduling of individual tasks. When several users share GPU clusters, the situation easily becomes complicated. One may easily imagine a typcial scenario where every user wants to use more resources and complete their tasks as quickly as possible.

The computing resources not only include GPUs, but also storage, memory, bandwidth, software, and even energy. Cloud computing, grid computing, IaaS, PaaS, and CaaS are the concepts which are usually presented to solve this  problem. Task-based scheduling can meet the demand for resources management of deep learning when the GPU, CPU, memory, and disk are considered as the main resources.

The management of resources usually includes task management and device management, as shown in Fig.\ \ref{fig:vispg}. The task management takes charge of the scheduling of tasks, while the device manager is in charge of the controlling and organizing of the hardware.

\paragraph{Task management} MeDaS employs Docker and Kubernetes to manage containers. Docker containers use the ``control group'' to establish a sandbox with limited devices allocated. Each task has all the resources to itself in a container. The tasks are scheduled with containers.

\paragraph{Device management} The number of GPUs is controlled with a different set of the plan according to the calculation scale. The device management is controlled by using Docker and Kubernetes.

%% file: application.tex

\begin{figure*}
    \centering
    \includegraphics[width=\linewidth]{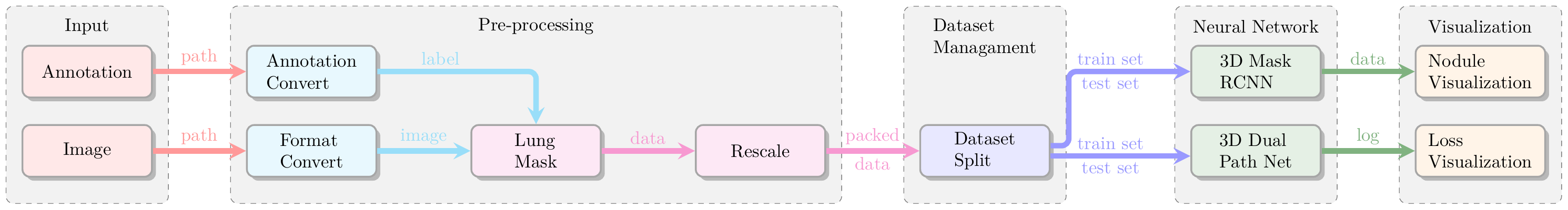}
    \caption{Workflow and data flow of the pulmonary nodule detection and the attribute classification (case study \ref{case:pndc}).
        The workflow includes five parts: ``input'', ``pre-processing'', ``dataset management'', ``neural network'', and ``visualization''. A 3D mask RCNN is employed to detect, while a 3D dual-path net is employed for attribute classification.}
    \label{fig:workflow-demo1}
\end{figure*}

\begin{figure}
    \centering
    \subfloat{\includegraphics[width=0.5\linewidth]{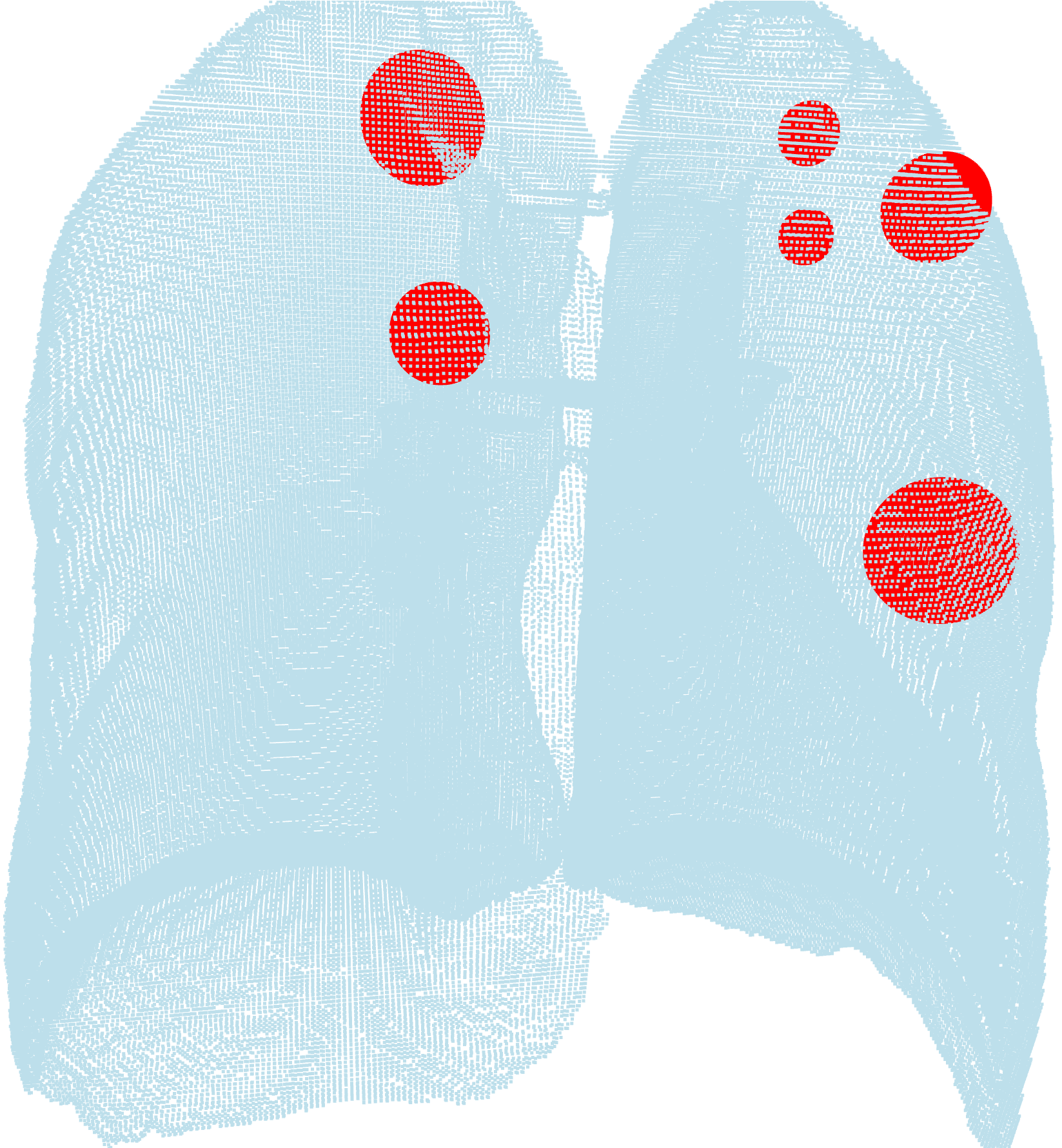}}
    \subfloat{\includegraphics[width=0.5\linewidth]{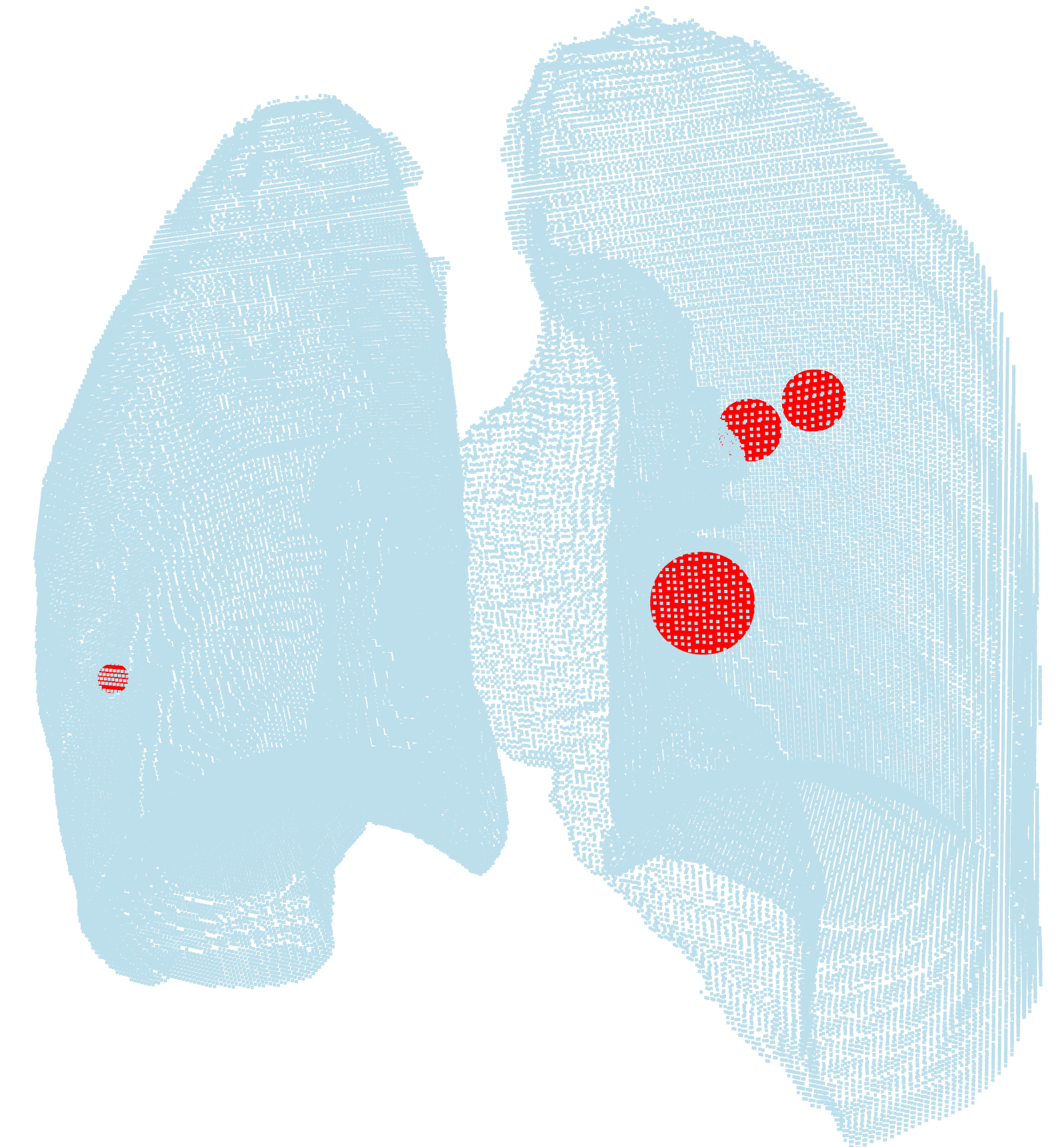}}
    \caption{The pulmonary nodules detected in two subjects. The red marks are the detected pulmonary nodules, while the blue points are the edges of the lung.}
    \label{fig:lung-nodules}
\end{figure}

\section{Application Case Studies}
\label{sec:app}

In this section, we present different case studies performed using MeDaS, and selected varying themes of tasks. We thereby focus on the different parts of MeDaS in these case studies. Deep learning-based methods are employed throughout in these case studies.
The case studies include:
\begin{itemize}
    \item[] \nameref{case:pndc}
    \item[] \nameref{case:lcs}
    \item[] \nameref{case:mos}
    \item[] \nameref{case:ad}
    \item[] \nameref{case:ns}
\end{itemize}

On purpose to foster comparability and reproducibility, we chose public datasets of medical image analysis tasks in these case studies.
Each case study will introduce the workflow of the model, and the workflow is implemented with visualization programming via simple drag and drop. The results of the model will be shown in each case study.

These case studies are executed with MeDaS via the premium container.\footnote{The container includes 6 cores of Intel\textregistered~Xeon\textregistered~Gold 5120, an NVIDIA Tesla V100(32G PCIE version), and 48 Gigabytes of memory.}

\newdemo{Pulmonary Nodule Detection \& Attribute Classification}
\label{case:pndc}

The detection and attribute classification of the pulmonary nodule is a common medical image analysis task and is important for lung cancer diagnosis and clinical treatment. In this case study, we employ the ``deep lung''-based neural network, proposed in \cite{Zhu2018}, to detect and classify the pulmonary nodule.

The LUNA 16 dataset, which is based on the LIDC-IDRI dataset, \cite{ArmatoIIISamuelG.2015}, is used to train the model. The details of this case study are provided in the following subsections.

\begin{figure}
    \centering
    \subfloat[total loss]{\includegraphics[width=.5\linewidth]{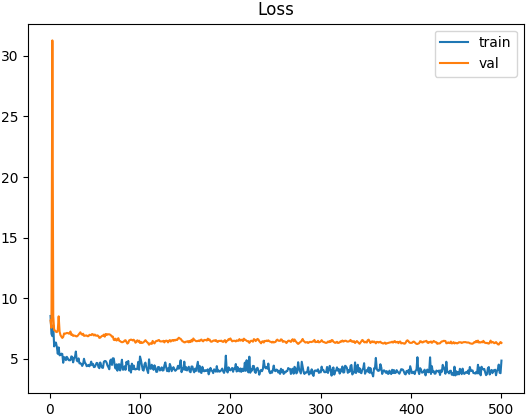}}
    \subfloat[classification loss]{\includegraphics[width=.5\linewidth]{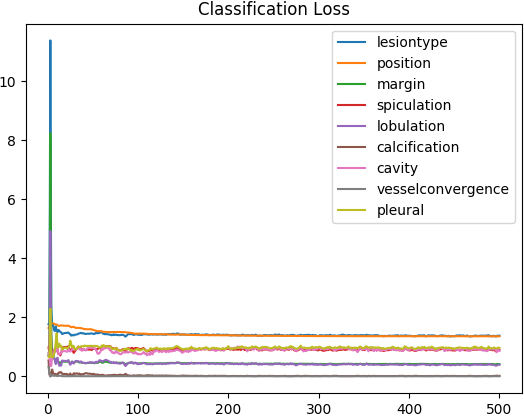}}
    \caption{The total loss and separate classification loss. The left plot shows the training loss (blue line) and testing loss (orange line). The right plot shows the loss of different classifiers.}
    \label{fig:demo1:loss}
\end{figure}

\subsubsection{Workflow}

\begin{figure*}
    \centering
    \includegraphics[width=\linewidth]{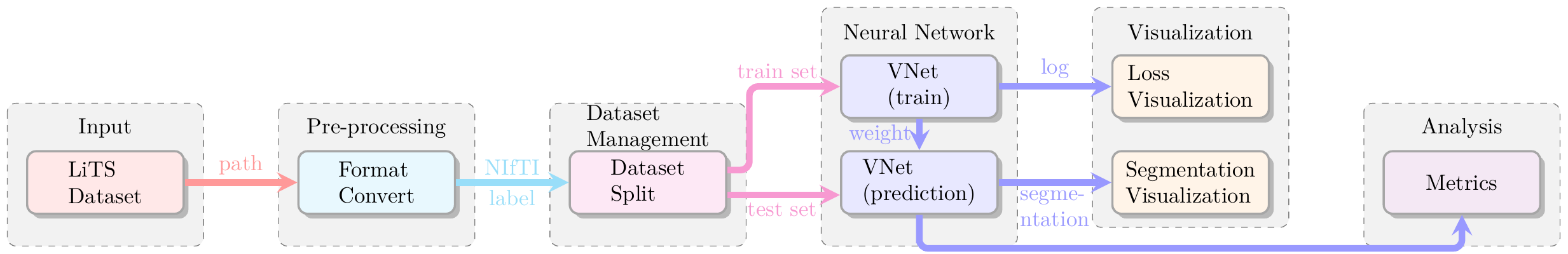}
    \caption{Workflow and data flow for case study \ref{case:lcs}. The workflow includes six parts: ``input'', ``pre-processing'', ``dataset management'', ``neural network'', ``visualization'', and ``analysis''.}
    \label{fig:workflow-demo2}
\end{figure*}

The basic workflow is shown in Fig.\ \ref{fig:workflow-demo1}.
The workflow is split into five parts, including ``input'', ``pre-processing'', ``dataset management'', ``neural network'', and ``visualization''.
The ``input'' is the source of data.

``Pre-processing'' tools  convert the formats of the image, annotate it, and process the data. The ``lung mask'' tool can  generate the mask of the lung, and help the deep learning model to focus on it and reduce the noise. The ``rescale'' tool can transform the values of inputs. The value of the image will be limited with the window width and window level of the lung, and rescaled to $0$ to $1$, which is a common range used in deep learning.

``Dataset management'' is used to split the dataset into a training set and a testing set. Note that some deep learning tasks might split the dataset into a training set, a validation set, and a testing set. The training set will be used to train the model, while the testing set or validation set can be used to evaluate the model when or after training.

``Neural Network'' employs 3D Mask RCNN for pulmonary nodule detection, while the 3D Dual-Path Net is used for attribute classification.

``Visualization'' employs point cloud-based nodule visualization to display the pulmonary nodule detected by the 3D mask RCNN, and the loss visualization can visualize the training loss of the model.

\subsubsection{Implementation}

Simple steps by dragging and dropping with MeDaS can implement the workflow mentioned in the previous. Then we can launch the Docker container and load data from a database to execute the task.

\subsubsection{Result and Visualization}

We train the 3D mask RCNN model and the 3D dual-path net with the training set and test them with the testing set. Fig.\ \ref{fig:lung-nodules}, which is rendered via the 3D point cloud, shows the result of the 3D mask RCNN. Fig.\ \ref{fig:demo1:loss} presents the training loss of the 3D dual-path net. The left plot shows the total loss, while the right plot presents the loss for every classifier.

\begin{figure*}
    \centering
    \includegraphics[width=\linewidth]{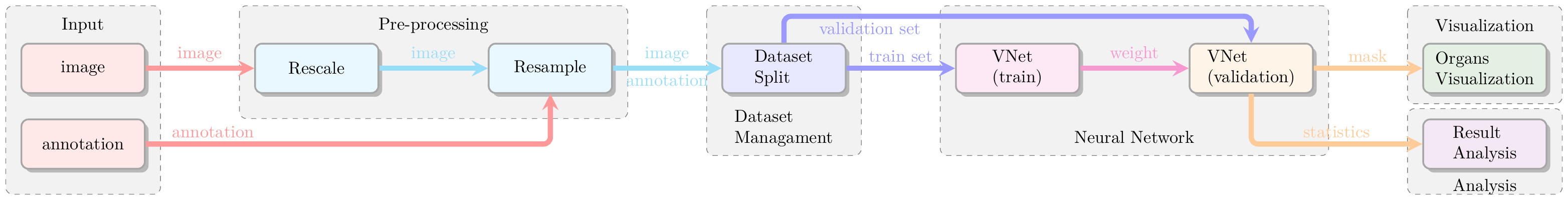}
    \caption{The workflow of multi-organ segmentation (case study \ref{case:mos}). The workflow includes the pre-processing of data and annotations, the training, the evaluation, and the visualization.}
    \label{fig:workflow-demo3}
\end{figure*}

\newdemo{Liver Contour Segmentation}
\label{case:lcs}

Deep learning-based methods are also a hot research direction in the liver-related radio-graphic analysis. The first step is usually to segment the contour of the liver. In this case study, we employ a VNet-based neural network, proposed in \cite{Milletari2016}, to segment liver contours.

The public dataset LiTS \cite{Bilic2019} is used to train the model. This dataset is aimed at the detection and segmentation of the liver and tumors.

\subsubsection{Workflow}
As shown in Fig.\ \ref{fig:workflow-demo2}, the workflow of this case study includes six parts. The ``input'' part is the source of data, and we use ``pre-processing'' to convert formats of images. Next, the dataset is split into a training set and a testing set.

The VNet is employed to segment the liver contours from the images and trained with the training set. Then, we use the trained model to initialize the prediction tool of the model to test the testing set. The training loss is visualized with the ``loss visualization'' tool, while the segmentation results are presented with the ``segmentation visualization'' tool. The prediction and ground truth are analyzed by computing the ``dice score''.

\subsubsection{Implementation}
The algorithm can be developed by using MeDaS's visualization programming. However, in this case study, we will show the alternative option available to users to program in MeDaS.
The set up, execution, and results checking with the training tool will be shown as an example.

To use the tool, there are four steps to follow:
\begin{enumerate}
    \item initializing instances
    \item setting up the tool
    \item executing the tool
    \item checking the results
\end{enumerate}
shown as the following codes:
\begin{lstlisting}[language=python]
tool = TrainVNet()
tool.set_params(new_ct_dir = new_ct_dir, new_seg_dir = new_seg_dir, save_module_path = save_module_path, save_loss_path = save_loss_path)
tool.run()
tool.with_succ(handler)
\end{lstlisting}

With ``continuous programming'', the codes above is equal to the below one:

\begin{lstlisting}[language=python]
tool = TrainVNet()
    .set_params(new_ct_dir = new_ct_dir, new_seg_dir = new_seg_dir, save_module_path = save_module_path, save_loss_path = save_loss_path)
    .run()
    .with_succ(handler)
\end{lstlisting}

\subsubsection{Result and Visualization}

The network for liver contour segmentation is trained on the LiTS dataset, and the model obtains 0.96 as ``dice score''. The dice on the LiTS dataset (testing set) reaches 0.92 as the best performance.

\begin{figure}
    \centering
    \subfloat[CT]{\includegraphics[width=0.33\linewidth]{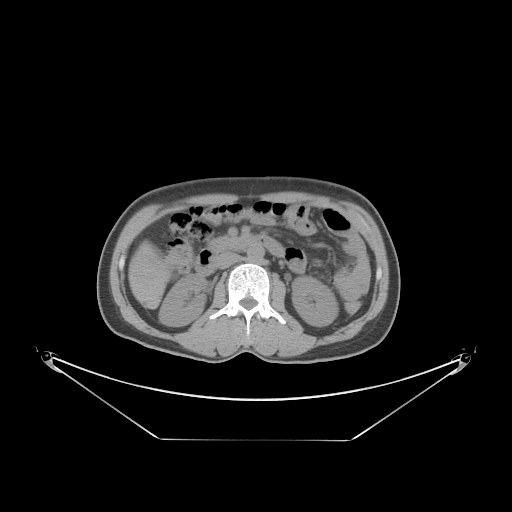}}
    \subfloat[Segmentation]{\includegraphics[width=0.33\linewidth]{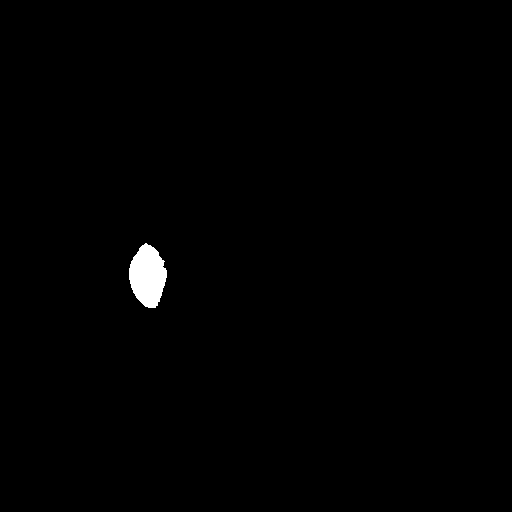}}
    \subfloat[Ground Truth]{\includegraphics[width=0.33\linewidth]{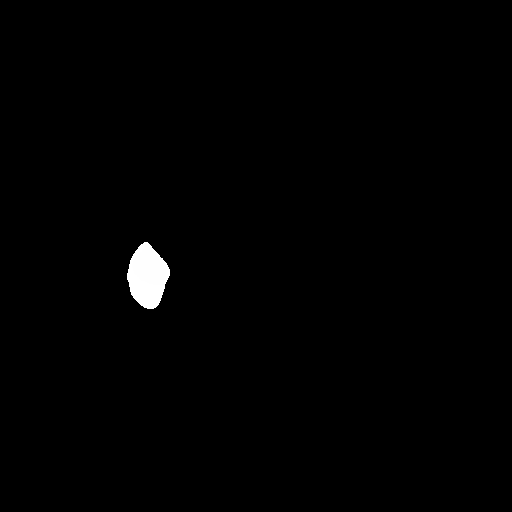}}\\
    \subfloat[CT]{\includegraphics[width=0.33\linewidth]{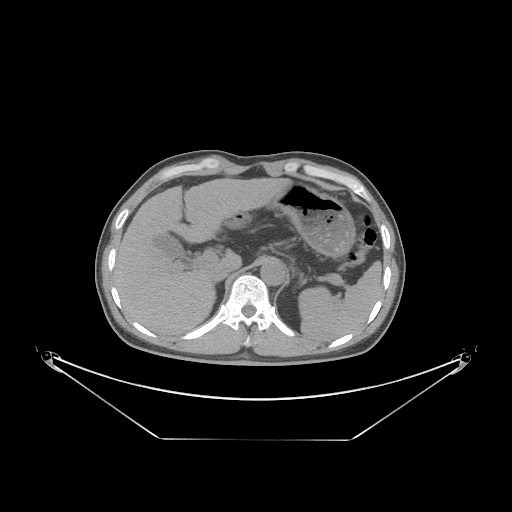}}
    \subfloat[Segmentation]{\includegraphics[width=0.33\linewidth]{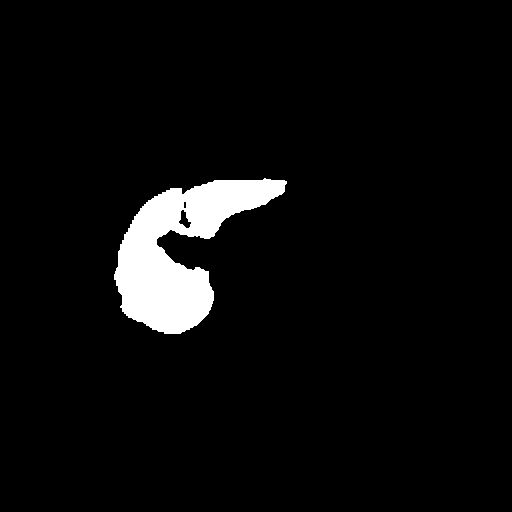}}
    \subfloat[Ground Truth]{\includegraphics[width=0.33\linewidth]{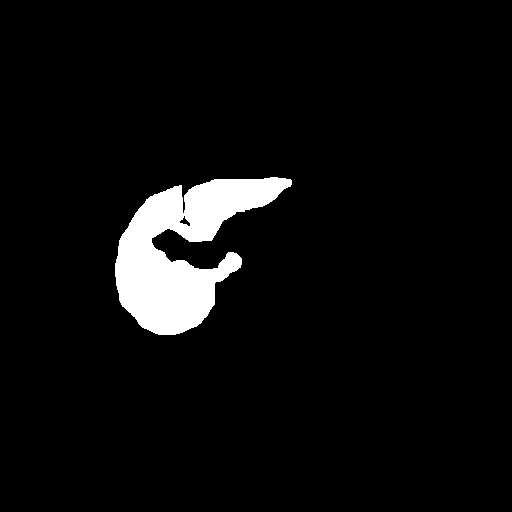}}\\
    \subfloat[CT]{\includegraphics[width=0.33\linewidth]{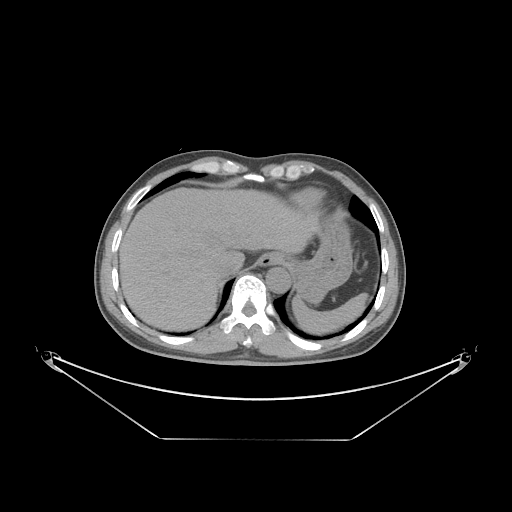}}
    \subfloat[Segmentation]{\includegraphics[width=0.33\linewidth]{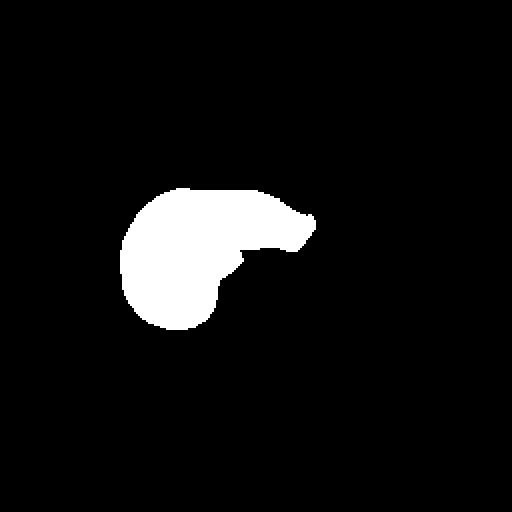}}
    \subfloat[Ground Truth]{\includegraphics[width=0.33\linewidth]{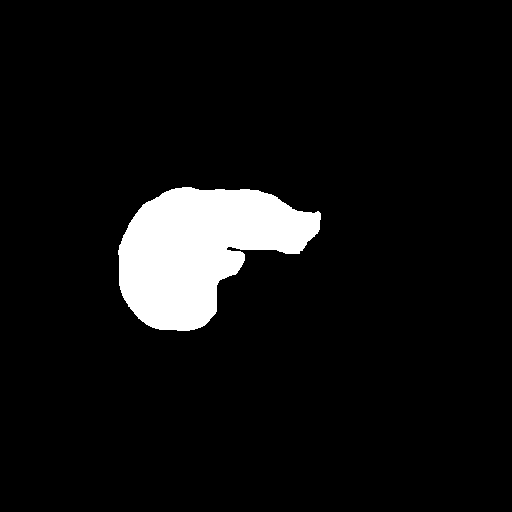}}
    \caption{The segmentation of the liver of three subjects. The left column is the original CT image, and the window width and level are 400 and 0. The middle image is the binary segmentations predicted by our model, while the right-most image is the ground truth.}
    \label{fig:demo2:result}
\end{figure}

\begin{figure}
    \centering
    \subfloat[Case study \ref{case:lcs}]{\includegraphics[width=.5192\linewidth]{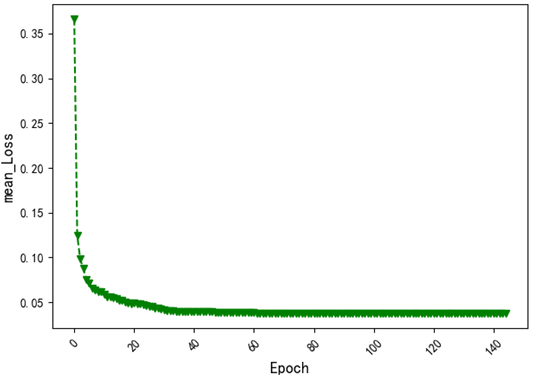}\label{fig:demo2:loss}}
    \subfloat[Case study \ref{case:mos}]{\includegraphics[width=.4808\linewidth]{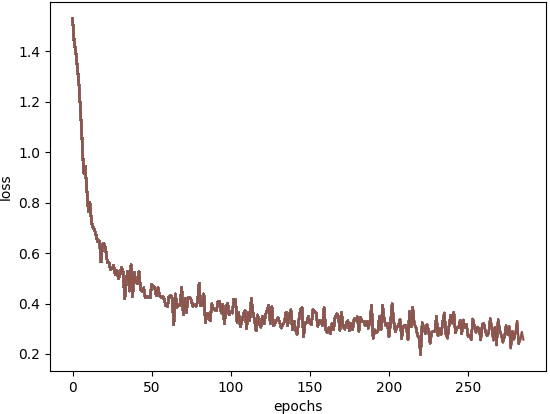}\label{fig:demo3:loss}}
    \caption{Visualization of the training loss for case study \ref{case:lcs} (left) and
        \ref{case:mos} (right).}
    \label{fig:demo:loss1}
\end{figure}

Fig.\ \ref{fig:demo2:result} presents the result of the segmentation task, while Fig.\ \ref{fig:demo2:loss} visualizes the training loss as the debugging information.

\newdemo{Multi-Organ Segmentation}
\label{case:mos}

The cognition of artificial intelligence is important for computer-aided  diagnostic. Multi-organ segmentation can help the machine understand the structure of the human body, which is very important for all the relevant tasks. Therefore, some research has focused on single- or multi-organ segmentation tasks, such as the liver(\cite{Dou2016,Lu2017}) and the pancreas(\cite{Cai2017,Zhou2017}). In this case study, we use a VNet-based neural network to solve the multi-organ segmentation challenge, SegTHOR, \cite{Trullo2017}. The SegTHOR challenge includes about 40 CT images of the chest, and aims at the segmentation tasks of the heart, aorta, trachea, esophagus, and further more.

\subsubsection{Workflow and Implementation}

As Fig.\ \ref{fig:workflow-demo3} shows, The workflow of this case study includes six parts: ``input'', ``pre-processing'', ``dataset management'', ``neural network'', ``visualization'', and ``analysis''. The ``input'' includes images of the chest and annotations.

``Pre-processing'' rescales the range of the image values with a window width and a window level. Then, the input images are re-sampled with the ``resample'' tool to change their size. The ``dataset management'' function subsequently splits the dataset into a training and a testing set randomly, yet reproducibly.

``Neural network'' employs VNet to train and validate the model, which can be used to segment organs from the chest. Then, the segmented images can be visualized  via the ``organ visualization'' option, and the results can be analyzed with the ``result analysis'' tool to generate an MS-Excel based report.

\subsubsection{Task Management}

When users set up and submit their tasks, MeDaS generates codes and files for each task, and launches a Docker container. When a task requires specific resources, such as the GPU, the scheduler will allocate or link the resources  to the container. When the limit of one account or the total of resources is  reached, the task will be failing or queued in line and wait for another chance to re-launch, when all the resources are ready.

This case study requires a number of computing resources, because even by using all the available computation resources, the whole tests can require months to finish. However, there may not always be tasks which require too many resources.
If every researcher or research group owns their own resources, the devices will be wasted when their users do not execute sufficient tasks that require the resources. MeDaS can manage scattered resources and use them efficiently with the same theory used by cloud services.

\subsubsection{Result and Visualization}
\begin{figure}
    \centering
    \subfloat[Ground Truth]{\includegraphics[width=0.33\linewidth]{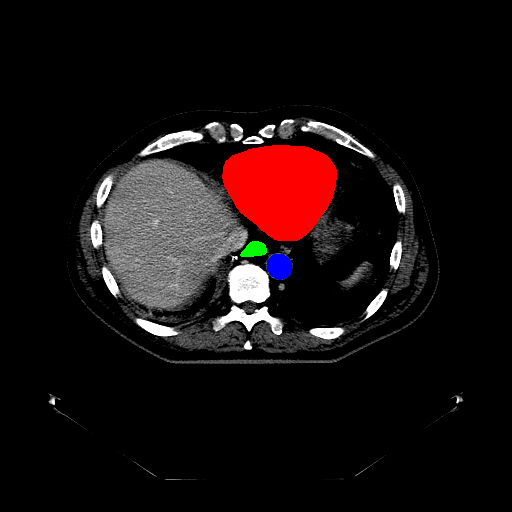}}
    \subfloat[Prediction]{\includegraphics[width=0.33\linewidth]{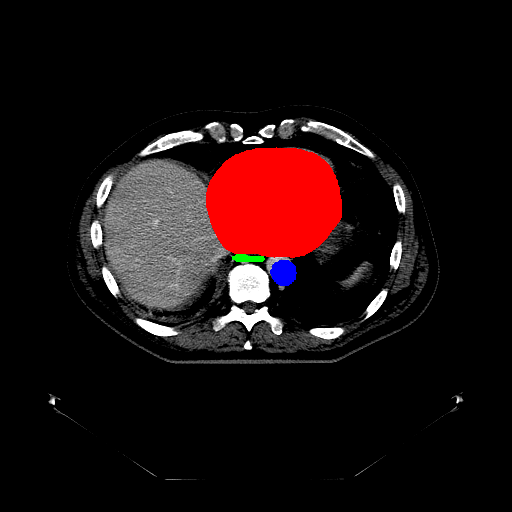}}
    \subfloat[Origin]{\includegraphics[width=0.33\linewidth]{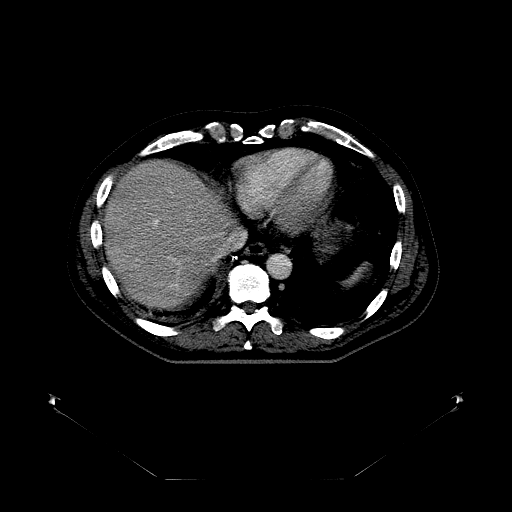}} \\

    \subfloat[Ground Truth]{\includegraphics[width=0.33\linewidth]{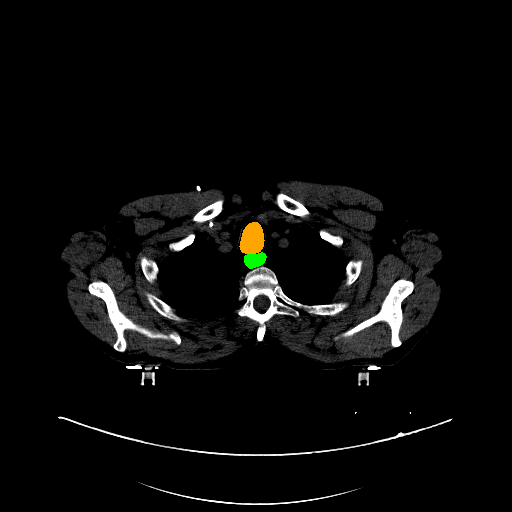}}
    \subfloat[Prediction]{\includegraphics[width=0.33\linewidth]{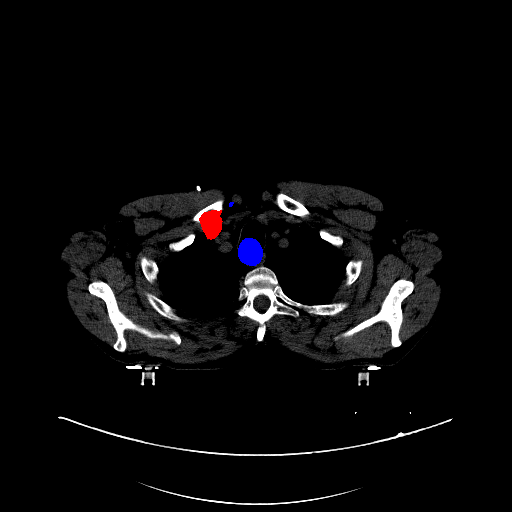}}
    \subfloat[Origin]{\includegraphics[width=0.33\linewidth]{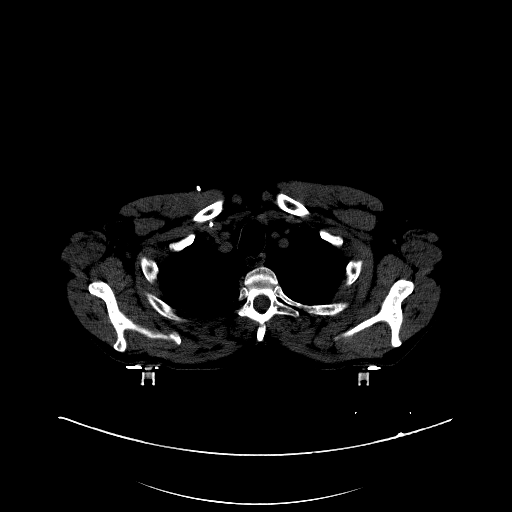}}
    \caption{Visualization of case study \ref{case:mos}. The green area is the esophagus; the red area is the heart; the blue area is the aorta; and the orange area is the trachea.}
    \label{fig:demo3:vis}
\end{figure}

Fig.\ \ref{fig:demo3:vis}  shows the obtained visualization results, and Fig.\ \ref{fig:demo3:loss} the training loss generated.

\begin{figure}
    \centering
    \includegraphics[width=\linewidth]{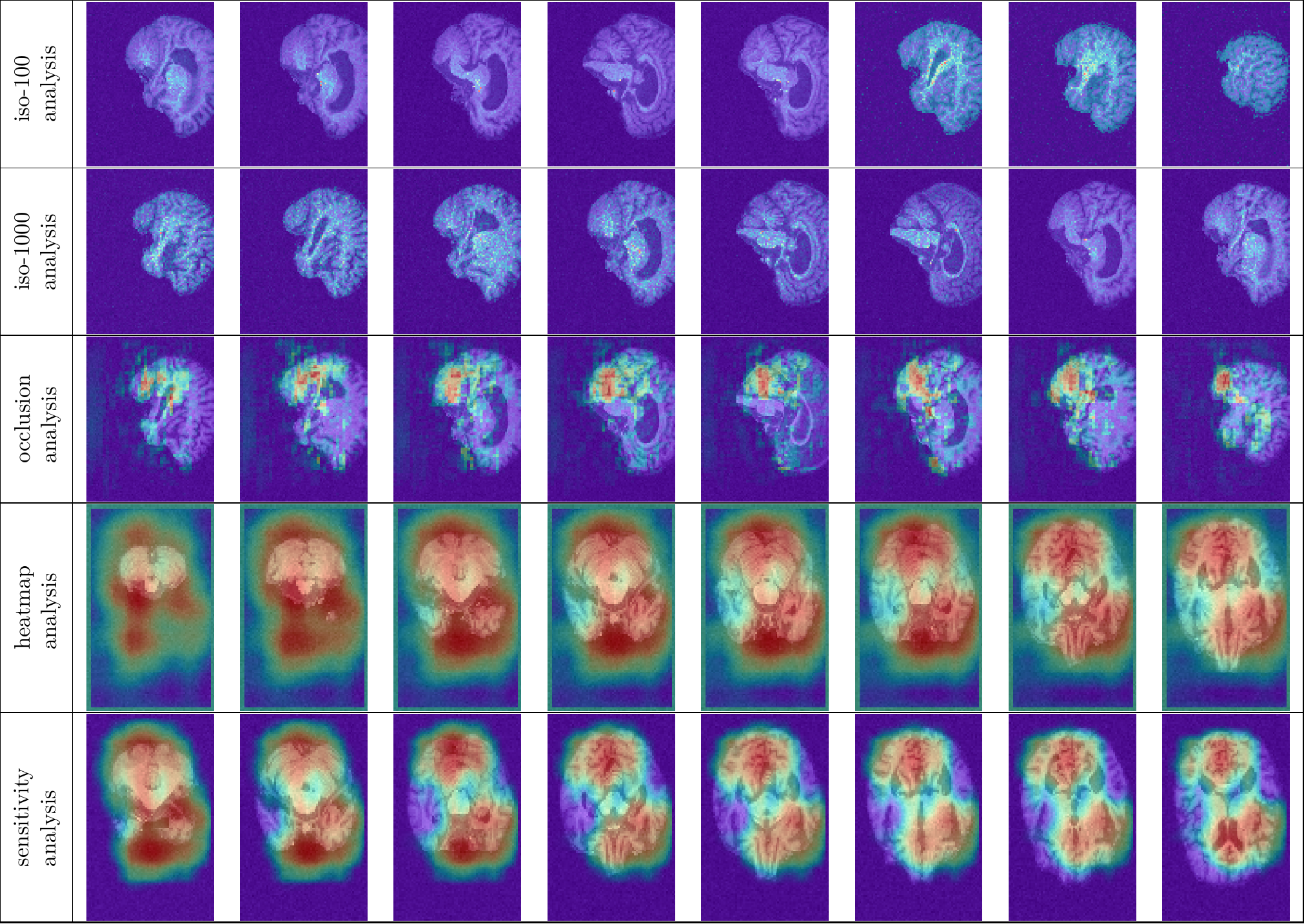}
    \caption{The heat map generated by the tool in MeDaS with block-based and contour-based occlusions. The first two rows resemble the analysis with color spacing split into different ranges. The third row includes the analysis results with occlusion. The fourth row resembles the activation heat-map. The last row depicts the sensitivity analysis result.}
    \label{fig:adheatmap}
\end{figure}

\newdemo{Alzheimer's Disease Classification}
\label{case:ad}

\begin{figure*}
    \centering
    \includegraphics[width=\linewidth]{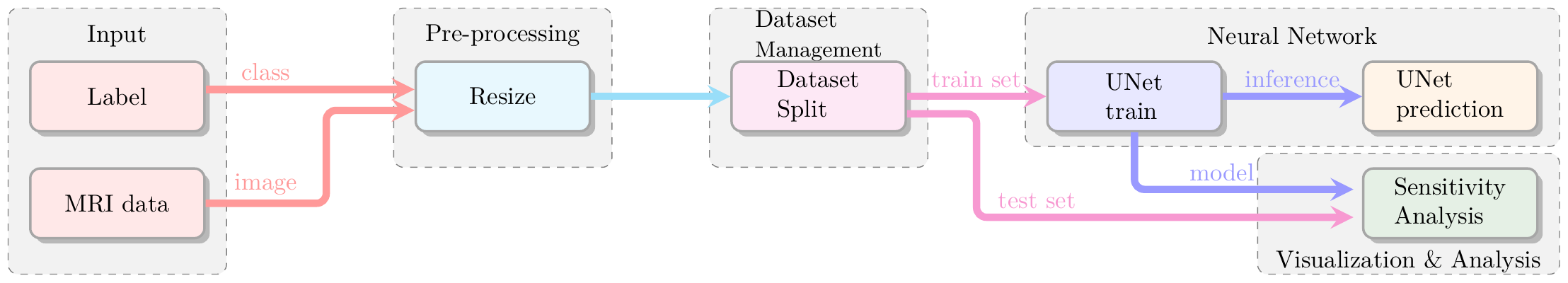}
    \caption{The workflow of Alzheimer's Disease Classification (case study \ref{case:ad}). The workflow includes the pre-processing of data and annotations, the training, the evaluation, and visualization.}
    \label{fig:workflow-demo4}
\end{figure*}

Alzheimer's disease (AD) is a kind of progressive neuro-degenerative disorder impairing the functions of memory and cognition according to \cite{Minati2009}.
Till now, there is no approach to cure the disease or even significantly slow down its deterioration, but there are some methods to tell the difference between AD and normal control (NC) cases, e.g., \cite{Khagi2019,Khvostikov2018,Khvostikov2017}. In this section, we employ U-Net, proposed in \cite{Ronneberger2015}, and modify it for classification tasks (AD versus NC).

In this case study, all the subjects are selected from a public AD dataset named ``the Alzheimer's Disease Neuroimaging Database'' (ADNI), \cite{Mueller2005}.  We select AD patients and NC data to train a classifier.

\subsubsection{Workflow}

As shown in Fig.\ \ref{fig:workflow-demo4}, the workflow of this case study includes five stages: input, pre-processing, dataset management, neural network, and visualization.

The ``input'' loads the data from the dataset, and the ``pre-processing'' tool processes data by resizing the images. The tool in the ``dataset management'' splits the dataset into two: a training set and a testing set.

The ``neural network'' includes a UNet-based classification network to filter AD from NC cases. The ``sensitivity analysis'' tool in ``visualization \& analysis'' helps to identify what is relevant for the neural network, by generating a heat map, which shows how the neural network behaves when a part of the image is occluded.

\subsubsection{Result}

The network is trained on MeDaS with the default parameters. A skip-connection version employs the AD vs NC classification.  The accuracy average for the classification task is 0.95.

\subsubsection{Visualization}

Generally, deep learning is considered as a black-box algorithm. It is difficult for researchers to understand what has been learnt by the neural network, and why the algorithm works well. For traditional algorithms, researchers have established models from clear reasons and targets, but for deep learning, only a general target is selected to let gradient descent optimize their models. A general neural network model for a complex task might include more than millions of parameters, which not only is hard to optimize,  but also troublesome to find out the effect of each parameter.

 MeDaS employs a lot of tools to help researchers visualize their models and results. In this case study, the results of the visualization are shown.
As shown in Fig.\ \ref{fig:adheatmap},  we employ three methods to visualize  the attention mechanism of our network. Such a tool can easily be used for similar tasks to generate heat map-based interpretable images. Block-based and contour-based occlusions are employed to interpret our model.

\begin{figure*}
    \centering
    \includegraphics[width=\linewidth]{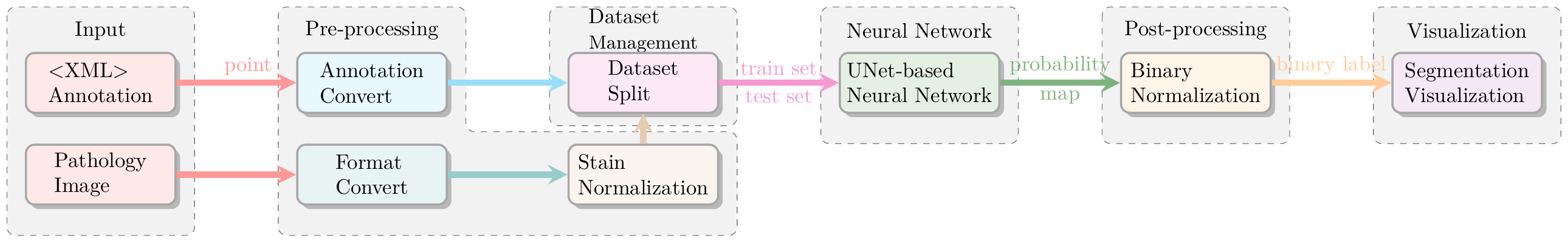}
    \caption{The workflow of Nuclei Segmentation (case study \ref{case:ns}). The workflow includes the pre-processing of data and annotations, the training, the evaluation, and the visualization.}
    \label{fig:workflow-demo6}
\end{figure*}

\newdemo{Nuclei Segmentation}
\label{case:ns}

Nuclei segmentation is one of the basic  pathology tasks in medical image analysis, whether it is traditional, \cite{Qaiser2019}, or deep learning-based, \cite{Song2019,Tofighi2019,Naylor2019}. The diagnostic of pathology images is based on many terms representing objects, such as nuclei, cells, glands, and so on. Researchers extract features from these objects and use them for higher-level diagnosis. For example, tasks such as mitosis analysis, depend on nuclei segmentation and detection, because the classification of mitosis is based on the nuclei.
We use a U-Net based model (\cite{Ronneberger2015})  to segment the nuclei on the dataset described in \cite{Kumar2017}.

\subsubsection{Workflow}

As shown in Fig.\ \ref{fig:workflow-demo6}, the workflow includes six stages.
The input loads the data from the dataset. The pre-processing converts formats and normalizes the stain of the pathology image. The dataset management splits the dataset into two sets, while the neural network uses the training set to train the model and uses the testing set to validate it.

The post-processing handles the results of the segmentation. It uses the binary normalization algorithm to improve the segmentation.
The visualization tool depicts the final results to the user.

\subsubsection{Implementation}

After the general design of the workflow, which can be done on the draft,
users can drop selected tools in the editor and connect them according to the data stream.

Then, the data is uploaded into the platform from a local-host or storage system, which is connected with the annotation information systems. Finally, the task and train the model are launched on MeDaS with the given workflow. Subsequently, the results and intermediate data are stored in the system.

\subsubsection{Hyper-parameter Optimization}

This case serves as an example of hyper-parameter optimization. Selected hyper-parameters in the neural network were picked carefully for optimization.

The range of the hyper-parameter ``max epoch'' is set to be chosen within 64 to 256, while the learning rate search range is set from 0.0001 to 0.01. The selection criteria are ``dice'', ``bce'', and ``lovasz'', while the selected models are FCN, ResUNet, DPUNet. At the same time, ``Mean AJI'' is selected as the optimization objective.

\begin{figure}
    \centering
    \subfloat{\includegraphics[width=\linewidth]{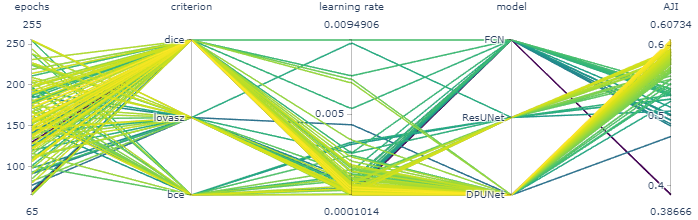}} \\
    \subfloat{\includegraphics[width=\linewidth]{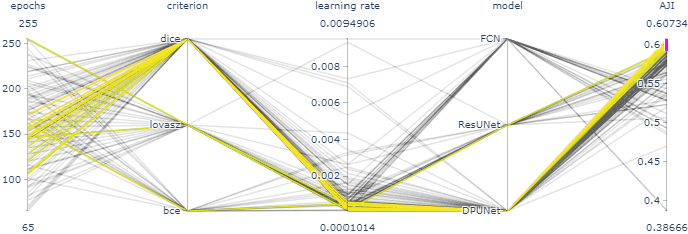}}
    \caption{The visualization of hyper-parameters via the parallel coordinates. The top one shows all the hyper-parameters. The color of the lines is related to the metric AJI: the higher, the brighter. The bottom one shows the hyper-parameters,  whose metric AJI ranges between 0.5925 to 0.6075.}
    \label{fig:demo6:ho}
\end{figure}

We performed 100 iterations to search with the Bayesian optimization algorithm.
The best result of the hyper-parameter optimization and the top five results of manual optimization are shown in Tables \ref{tab:demo6:ho} and \ref{tab:demo6:mo}. Further, Fig.\ \ref{fig:demo6:ho} shows the distribution of the hyper-parameters and the metric AJI. Most of the combinations with  ``DPUNet'' as a model and ``dice'' as a criterion function show better performance, which suggests that the metric AJI is higher, and the combinations, whose metric AJI is between 0.5925 to 0.6075, use them. From the figure, we can find that the epoch number of the training iterations does not result in a greater effect to the metric as compared to the criterion function, or say loss function, and the model. Further, the smaller learning rate proves the best choice , in general.

\begin{table}
    \centering
    \caption{The best results of the optimization. The max epoch, criterion, learning rate, number of training epochs, and model are selected as parameters.}
    \label{tab:demo6:ho}
\begin{tabular}{|c|c|c|c|c|c|}
    \hline
    epoch & criterion & learning rate         & model  & Mean AJI           \\ \hline
    172       & dice      & 4.081e-3  & DPUNet & 0.6073 \\ \hline
\end{tabular}
\end{table}

\begin{table}
    \centering
    \caption{The top-five results of a manual optimization with different parameters.}
    \label{tab:demo6:mo}
    \begin{tabular}{|c|c|c|c|c|}
        \hline
        epoch & criterion & learning rate & model   & Mean AJI           \\ \hline
        200       & dice      & 0.5e-3         & ResUNet & 0.5855 \\ \hline
        200       & dice      & 0.5e-3         & DPUNet  & 0.5854 \\ \hline
        256       & lovasz    & 0.25e-3        & ResUNet & 0.5832 \\ \hline
        128       & bce       & 1.0e-3         & DPUNet  & 0.5828 \\ \hline
        500       & lovasz    & 1.0e-3         & FCN     & 0.5821 \\ \hline
    \end{tabular}
\end{table}

Table \ref{tab:demo6:mo} shows the manual optimization result as a comparison. When we try to optimize these hyper-parameters manually, we are usually facing several problems.

\textbf{The most important one} is how to optimize the parameters as it is difficult to find an analytical solution. As outlined, MeDaS employs the Bayesian optimization algorithm aiming to find optimal hyper-parameters.

\textbf{The second problem} is the time. Manual optimization needs a lot of time. After we launch the task, we need to wait for the task to finish to test another set of parameters. If we have executed a task,  we cannot launch another task after the latest one has finished, because we cannot estimate when the task will finish exactly.

\textbf{The third problem} is the resource. Manual optimization usually needs more resources to reach a good result, since it tends to be slower and inefficient.

\subsubsection{Result and Visualization}

\begin{figure}
    \centering
    \subfloat[Origianl image]{\includegraphics[width=\linewidth]{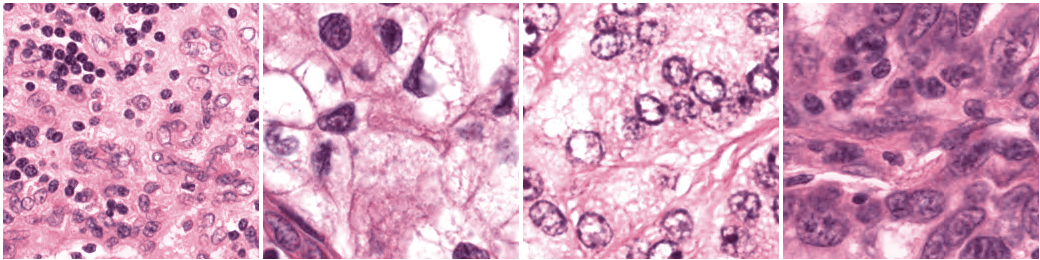}} \\
    \subfloat[Segmentation]{\includegraphics[width=\linewidth]{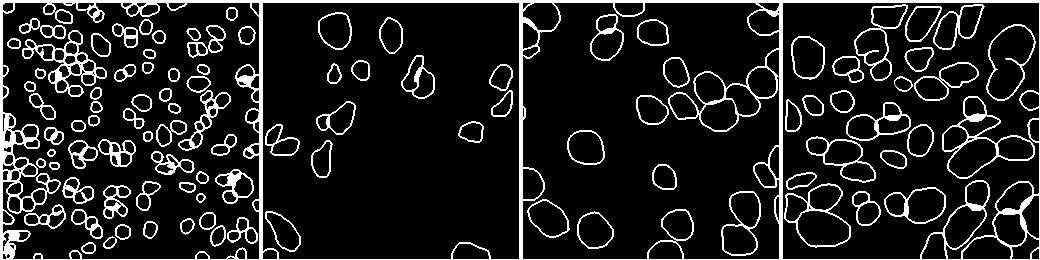}} \\
    \subfloat[Ground truth]{\includegraphics[width=\linewidth]{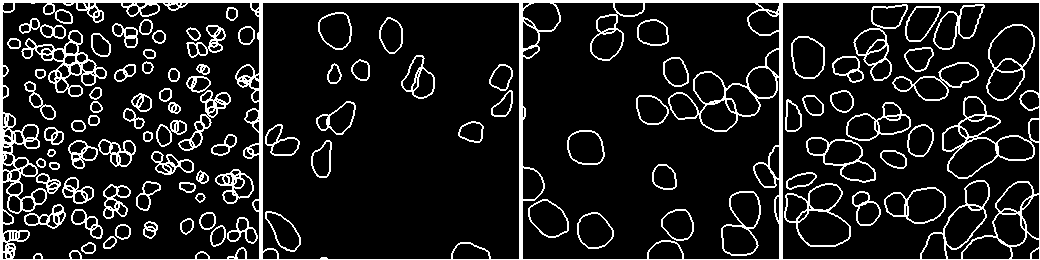}}
    \caption{Result of case study \ref{case:ns}. Each column shows the results of four different organs. The top row is the original image with pre-processing. The middle one is the segmentation with post-processing by binary normalization, while the bottom row is the ground truth.}
    \label{fig:case:6}
\end{figure}

The best result of the hyper-parameters is chosen as the final result. The DPUnet network is used as the model, and the dice function is selected as the loss function. The model is trained using 172 epochs with $4.081\times 10^{-4}$ as the learning rate. The metric, namely AJI, reaches 0.6073, and the segmentation of the nuclei is shown in Fig.\ \ref{fig:case:6}, while the AJI metrics of different organs are shown in Table \ref{tab:demo6:rs}.

\begin{table}
    \centering
    \caption{The AJI metric of the validation set for different organs.}
    \label{tab:demo6:rs}
    \begin{tabular}{|l|l|l|l|l|}
        \hline
        Organ & Breast & Liver & Bladder & Colon \\ \hline
        AJI & 0.6517 & 0.5310 & 0.6543 & 0.5424 \\ \hline
        Organ & Prostate & Stomach & Kidney & Mean \\ \hline
        AJI & 0.6147 & 0.6437 & 0.6135 &  0.6073 \\ \hline
    \end{tabular}
\end{table}

%% file: discussions.tex

\section{Discussion}
\label{sec:dscs}

\subsection{Shortcomings}

Aiming at the fusion of deep learning and medical image analysis, MeDaS provides a lot of functions that we though are useful. However, there are a lot of shortcomings in MeDaS. Firstly, MeDaS did not know well of its target user at the beginning of designing. Secondly, MeDaS lacks the ``strategy'' of small datasets. Thirdly MeDaS is not very industrial.

\subsubsection{Target Users}

Medical researchers are the target users, and they are more familiar with the software such as Microsoft Excel and IBM SPSS, compared with PyTorch or TensorFlow. MeDaS is 25\%  ``Excel'' and 75\%  ``PyTorch'', but medical researchers may need 25\%  ``PyTorch'' and 75\%  ``Excel''. Using simplest operations complete a relatively complex task is what they want. With or without visualization programming, medical researchers, till now, need to understand what are the neural network, the criterion function,  the optimization problem, the hyper-parameter search, and so on. Therefore, a well designed interface and automatic deep learning are future improvements.

\subsubsection{Small Datasets}
The datasets of medical image researches are usually in a small size, but deep learning needs the large size dataset.  MeDaS did not provide any components or functions to help researchers to use images crossing regions, such as hospitals. Distributed machine learning is one solution of such a problem, and it will be discussed in the next sub-section.

\subsubsection{Non-Industrial}
MeDaS is not ``very industrial''. Because MeDaS was born in a research environment, MeDaS was less designed for a application in the industrial context. The model cannnot be easily deployed to a production server after it was trained in the development server, MeDaS.

\subsection{Outlook}

The combination of deep learning and medical image analysis will still be a hot topic in the next few years, and a key problem in the present context is to fill the knowledge gap between medicine, deep learning, and computer science. The innovation of accessible technologies, like MeDaS, and methods will help  the progress in this direction.

\subsubsection{Automatic Deep Learning in Medical Informatics}

Automatic deep learning can help researchers to automatically design models and search for the best hyper-parameters. Hyper-parameter optimization and neural network architecture search are the problems that deep learning researchers need to face. This comes, as the choice of hyper-parameters and the design of the neural networks does not follow any specific rules.

Luckily, it is by now possible with automatic deep learning that medical researchers can simply input their data into the system, and it can design and search the best model for the tasks.

\subsubsection{Knowledge}

Medical knowledge can help on the deep learning end to understand what the machine has learnt, reach a medical explanation, and improve the methods, while deep learning can help to extract features ignored by humans.

Medicine and deep learning concerned surveys, papers, and even blog posts can be collected as a kind of knowledge base. For medical researchers, they can quickly find deep learning related knowledge which is used in their research, and for deep learning researchers, they can also rapidly retrieve medical knowledge. With MeDaS and such a knowledge base, both deep learning and medical researchers can accelerate their research.

\subsubsection{Federal learning and decentralized learning}

One difference between general computer vision and medical image analysis in deep learning is that the latter usually lacks data. First, most datasets are on a small scale. Comparing with many other computer vision datasets, such as SUN, most medical datasets only include tens or hundreds of subjects. Second, each group or laboratory might have their private datasets, but mostly on a small scale. These small isolated datasets make it difficult to use them.

Federal learning \cite{Sheller2019,BrendanMcMahan2017,Konecny2016} or other decentralized learning can help share what machines have learnt, but not sharing the data themselves. Based on platforms such as MeDaS, and decentralized learning, such as federal learning, researchers from different institutions can efficiently collaborate.

\subsubsection{Medical Image Attack and Defense}

The robustness analysis of an algorithm is an important step in medical image analysis. Researchers usually use statistic inspection or sensitive analysis methods to evaluate their algorithms. Medical image attacks and defense provide another approach of robustness evaluation. Finlayson et al. demonstrated there are vulnerabilities in deep learning, which might be used to crack deep learning-based medical image analysis system in their paper\cite{Finlayson2018}.

Medical image attacks and defense using GANs\cite{Mirsky2019} or similar methods to generate fake images, which humans cannot distinguish from real ones, can be used to attack and fool an algorithm and change the prediction of the algorithms. Hence, such a method can be used by medical researchers to evaluate their algorithms. Moreover, such an attack and defense can help to improve the  model's robustness. The fake images, which are used in the attack, can help the robustness of algorithms. After the algorithm has learnt the features in fake images with the right labels, the algorithm should be able to defend against the attack.

%% file: summary.tex

\section{Summary}
\label{sec:sum}

In this work, we introduced our platform, named MeDaS, to render the application of deep learning in the medical image analysis more user-friendly, easy, and hence accessible. We designed the pipeline and user interface based on our experience of development and analysis. The pipeline includes pre-processing, post-processing, augmentation, neural network modules, and visualization and debugging tools. We have also performed several case studies to demonstrate the efficient operation of MeDaS.

\section*{Acknowledgments}

The authors would like to acknowledge all of the contributors to \textit{MeDaS: An open-source platform as service to help break the walls between medicine and informatics}. This work was supported by Shanghai Science and Technology Committee (No.\ 18411952100, No.\ 17411953500)